\documentclass{article}

\usepackage{arxiv}

\usepackage[utf8]{inputenc} 
\usepackage{hyperref}       
\usepackage{lineno}
\usepackage{url}            
\usepackage{booktabs}       
\usepackage{amsfonts}       
\usepackage{amsmath}
\usepackage{nicefrac}       
\usepackage{microtype}      
\usepackage{lipsum}		
\usepackage{graphicx}
\usepackage[numbers]{natbib}
\usepackage{doi}
\usepackage{bm}
\usepackage{multirow} 
\usepackage{xcolor}
\usepackage{inconsolata}

\title{GENERator: A Long-Context Generative Genomic Foundation Model}

\author{
\textbf{Wei Wu}\textsuperscript{*,1,4}, 
\textbf{Qiuyi Li}\textsuperscript{*,{\dag},1,2,3}, 
\textbf{Yuanyuan Zhang}\textsuperscript{*,5,6}, 
\textbf{Zhihao Zhan}\textsuperscript{9,10}, 
\textbf{Ruipu Chen}\textsuperscript{5,6}, 
\textbf{Mingyang Li}\textsuperscript{1}, 
\\ \vskip 0.03in
\textbf{Kun Fu}\textsuperscript{1}, 
\textbf{Junyan Qi}\textsuperscript{5,6}, 
\textbf{Yongzhou Bao}\textsuperscript{5,6}, 
\textbf{Chao Wang}\textsuperscript{5,6}, 
\textbf{Yiheng Zhu}\textsuperscript{2,3}, 
\textbf{Zhiyun Zhang}\textsuperscript{13}, 
\\ \vskip 0.03in
\textbf{Jian Tang}\textsuperscript{9,11,12}, 
\textbf{Fuli Feng}\textsuperscript{4}, 
\textbf{Jieping Ye}\textsuperscript{1}, 
\textbf{Yuwen Liu}\textsuperscript{{\dag},5,6}, 
\textbf{Hui Xiong}\textsuperscript{{\dag},7,8}, 
\textbf{Zheng Wang}\textsuperscript{{\dag},1}}

\affiliation{
\textsuperscript{*}Equal Contribution \hspace{12pt} \textsuperscript{\dag}Equal Senior Authorship\\
\vskip 0.15in
\textsuperscript{1}Alibaba Cloud Computing, Beijing, China\\
\vskip 0.05in
\textsuperscript{2}Zhongguancun Academy, Beijing, China\\
\vskip 0.05in
\textsuperscript{3}Zhongguancun Institute of Artificial Intelligence, Beijing, China\\
\vskip 0.05in
\textsuperscript{4}University of Science and Technology of China, Hefei, China\\
\vskip 0.05in
\textsuperscript{5}State Key Laboratory of Genome and Multi-omics Technologies, Shenzhen Branch, Guangdong Laboratory for Lingnan Modern Agriculture, Key Laboratory of Livestock and Poultry Multi-Omics of MARA, Agricultural Genomics Institute at Shenzhen, Chinese Academy of Agricultural Sciences, Shenzhen, China\\
\vskip 0.05in
\textsuperscript{6}Innovation Group of Pig Genome Design and Breeding, Research Centre for Animal Genome, Agricultural Genomics Institute at Shenzhen, Chinese Academy of Agricultural Sciences, Shenzhen, China\\
\vskip 0.05in
\textsuperscript{7}The Hong Kong University of Science and Technology (Guangzhou), Guangzhou, China\\
\vskip 0.05in
\textsuperscript{8}The Hong Kong University of Science and Technology
, Hong Kong SAR, China\\
\vskip 0.05in
\textsuperscript{9}Mila - Qu\'ebec AI Institute, Montr\'eal, Canada\\
\vskip 0.05in
\textsuperscript{10}University of Montr\'eal, Montr\'eal, Canada\\
\vskip 0.05in
\textsuperscript{11}HEC Montr\'eal, Montr\'eal, Canada\\
\vskip 0.05in
\textsuperscript{12}CIFAR AI Chair, Canada\\
\vskip 0.05in
\textsuperscript{13}Carnegie Mellon University, Pittsburgh, USA\\
\vskip 0.15in
\textsuperscript{\dag}Correspondence to: \textit{qiuyi.li1993@gmail.com, liuyuwen@caas.cn, xionghui@ust.hk, wz388779@alibaba-inc.com}
}

\date{}

\begin{document}
\maketitle

\begin{abstract}
The rapid advancement of DNA sequencing has produced vast genomic datasets, yet the interpretation and rational engineering of sequence function remain fundamental challenges. Recent large language models (LLMs) have opened new avenues for genomic analysis; however, existing approaches are frequently constrained by limited training scope, restricted generative flexibility, or prohibitive computational cost. In this study, we introduce GENERator, a generative genomic foundation model designed for long-context DNA modeling, with a context length of 98k nucleotides, pre-trained on 386 billion nucleotides of eukaryotic DNA.
GENERator demonstrates strong intrinsic capabilities arising directly from pre-training. Unsupervised embedding analyses reveal latent organization consistent with phylogenetic relationships. Sequence recovery benchmarks show that GENERator achieves generative accuracy matching or exceeding state-of-the-art baselines with substantially improved computational efficiency. In a zero-shot setting, GENERator further provides competitive variant effect prediction performance relative to alignment-based methods, while remaining fully alignment-free and broadly applicable across species. Beyond training-free evaluation, GENERator consistently delivers strong performance through task-specific fine-tuning on established genomic benchmarks.
We further demonstrate practical generative applications enabled by the model. GENERator can generate protein-coding DNA sequences that translate into structurally plausible proteins and, through a prompt-guided design framework, design cis-regulatory elements with targeted activity profiles, including synthetic enhancers whose regulatory strength exceeds that of natural genomic sequences, as validated by high-throughput UMI-STARR-seq assays. Collectively, these results establish GENERator as an efficient and biologically grounded foundation for genomic interpretation and programmable sequence design across diverse genomic contexts. Implementation details and supplementary resources are available at \url{https://github.com/GenerTeam/GENERator}.
\end{abstract}

\section{Introduction}

Genomic sequences encode rich biological information that underlies diverse molecular processes, ranging from gene regulation to protein synthesis, and ultimately shapes phenotypic variation and disease susceptibility~\cite{nucleotide-sequence,bioinformatics-and-functional-genomics,civelek2014systems}. Advances in DNA sequencing technologies~\cite{next-generation-dna-sequencing} have enabled large-scale decoding of genomes across a broad range of organisms. Despite this progress, reliably interpreting and engineering genomic function remains a major challenge, owing to the complex organization of genetic material and the limited availability of high-quality, functionally annotated training data.

Recent advances in machine learning, particularly large language models (LLMs)~\cite{llms-survey}, have opened new avenues for biological sequence analysis. In natural language processing, models such as BERT~\cite{BERT} and GPT~\cite{GPT4} have demonstrated the power of large-scale pre-training for transferable representation learning. Similar principles have driven progress in biological domains, exemplified by AlphaFold~\cite{AlphaFold2,AlphaFold3} and ESM~\cite{esm2,esm3} for protein structure and function modeling. In genomics, early foundation models have predominantly adopted masked language modeling (MLM) objectives, including DNABERT~\cite{DNABERT,DNABERT-2}, LucaOne~\cite{LucaOne}, GROVER~\cite{GROVER}, Caduceus~\cite{Caduceus}, and the Nucleotide Transformer (NT)~\cite{nucleotide-transformer}. While these models excel at learning contextual representations of DNA sequences, they are not naturally optimized for open-ended sequence generation or functional design and are typically limited to relatively short context lengths.

Generative modeling offers a complementary paradigm by unifying sequence understanding with the ability to synthesize novel genomic content. Autoregressive causal language models, such as HyenaDNA~\cite{HyenaDNA}, megaDNA~\cite{megaDNA}, Evo~\cite{Evo} and GenomeOcean~\cite{genomeocean}, enable direct sequence generation through next-token prediction, while diffusion-based approaches~\cite{D3, MDLM, DDSM} model sequence distributions via iterative denoising. However, most existing generative genomic models are trained on restricted domains, such as human genomes~\cite{HyenaDNA}, bacteriophages~\cite{megaDNA}, or bacterial and viral sequences~\cite{Evo, genomeocean}, limiting their applicability to the broader eukaryotic landscape characterized by complex gene architectures and regulatory mechanisms. 

The recently released Evo2~\cite{Evo2} represents a major step forward in generative genomic modeling by jointly modeling prokaryotic, eukaryotic, and viral genomes at unprecedented scale. This unified training strategy substantially expands biological coverage, but also incurs significant computational cost, which presents practical challenges for training, inference, and downstream experimentation. These considerations motivate the development of generative genomic foundation models that can deliver strong biological performance while operating under more accessible and efficient computational regimes.

\begin{figure*}[t]
    \centering
    \includegraphics[width=0.9\textwidth]{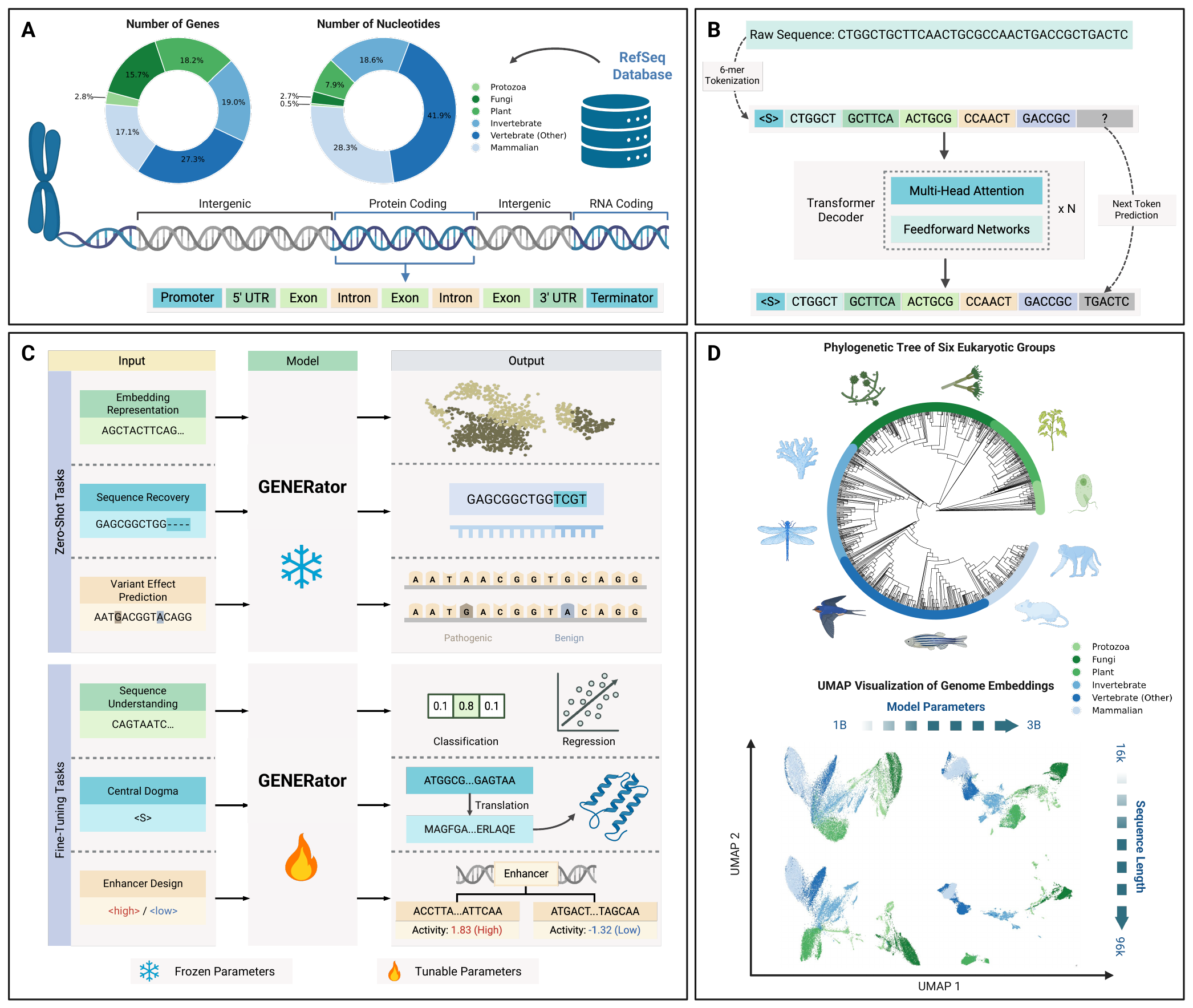}
    \caption{Overview of the GENERator model.
    (A) GENERator is pretrained on large-scale eukaryotic genomic sequences from the RefSeq database, spanning all major eukaryotic lineages. We adopt a functional sequence training strategy that leverages RefSeq annotations to extract gene-centric functional regions, and perform pretraining exclusively on these biologically meaningful sequences.
    (B) GENERator is a generative DNA language model based on a transformer decoder architecture with 6-mer tokenization, pretrained using an autoregressive next-token prediction objective to learn sequence dependencies and regulatory grammars from functional genomic data.
    (C) GENERator enables both training-free (zero-shot) and task-specific fine-tuning applications. Zero-shot tasks include genomic embedding representation, prompt-conditioned sequence recovery and generation, and variant effect prediction for benign versus pathogenic mutations. Fine-tuned tasks include supervised sequence classification and regression (e.g., promoter classification and enhancer activity prediction), central-dogma tasks involving protein-coding sequence generation and translation, and prompt-responsive cis-regulatory element (CRE) design for generating enhancers with specified activity profiles.
    (D) Top: phylogenetic tree illustrating the eukaryotic species represented in the RefSeq dataset. Bottom: UMAP visualizations of genome embeddings learned by GENERator, showing taxonomically consistent clustering. Increasing sequence length (16k to 96k) and model size (1B to 3B parameters) leads to progressively improved embedding separation, indicating clear scaling effects in representation quality.}
    \label{fig:model_overview}
\end{figure*}

In this work, we introduce GENERator, a generative genomic foundation model built upon an autoregressive DNA language modeling framework based on a transformer decoder architecture. The model is pre-trained on 386 billion nucleotides of eukaryotic DNA curated from the RefSeq database~\cite{RefSeq}. To balance sequence resolution with effective context coverage under realistic computational budgets, GENERator adopts a 6-mer tokenization strategy, which reduces the effective token sequence length, enabling the modeling of substantially longer genomic contexts. Combined with an autoregressive training objective, this design supports expressive sequence modeling and flexible generative applications. An overview of the GENERator framework is shown in Figure~\ref{fig:model_overview}.

The scale and phylogenetic diversity of the training data endow GENERator with robust representational and generative capacity across diverse organisms. We first assess intrinsic, task-agnostic properties that emerge directly from pre-training. Unsupervised embedding analyses show that genomic sequences are organized in latent space according to deep evolutionary relationships. To evaluate generative fidelity in a unified manner, we introduce a sequence recovery task that enables fair comparison across tokenization strategies and model architectures. Under this evaluation, GENERator achieves generative accuracy comparable to or exceeding strong state-of-the-art baselines, including Evo2, while operating with substantially improved computational efficiency.  

In addition, GENERator supports alignment-free variant effect prediction in a zero-shot setting, delivering competitive performance relative to alignment-based methods such as GPN-MSA~\cite{gpn-msa} and CADD~\cite{cadd}, while remaining readily applicable across species.

Beyond training-free evaluation, GENERator demonstrates strong adaptability through task-specific fine-tuning, achieving leading performance across established genomic benchmarks, including NT tasks~\cite{nucleotide-transformer}, Genomic Benchmarks~\cite{genomic-benchmarks}, and the newly proposed Gener tasks. We further demonstrate its utility in biologically grounded generative applications across multiple biological scales. Consistent with the central dogma of molecular biology~\cite{central-dogma}, GENERator generates protein-coding DNA sequences that translate into structurally plausible proteins. Moreover, through a prompt-guided design framework, the model enables programmable cis-regulatory element (CRE) design, generating regulatory sequences with targeted activity profiles, including synthetic enhancers whose regulatory strength exceeds that observed in natural counterparts.

Together, these results establish GENERator as a versatile, efficient, and biologically grounded genomic foundation model. By integrating scalable sequence modeling with generative design capabilities, GENERator provides a principled framework for interpreting, predicting, and engineering genomic sequences, opening new opportunities for functional genomics and synthetic biology.

\section{Results}
\label{sec:experiments}
\subsection{Embedding Clustering}
To assess the intrinsic representational quality of GENERator, we conducted unsupervised embedding clustering as described in Section~\ref{sec:embedding_clustering}. We constructed a diverse evaluation dataset by sampling fixed-length genomic segments from six eukaryotic taxonomic groups represented in RefSeq. To systematically examine the effects of model scale and input context, we evaluated four model configurations, combining two model scales (1B and 3B parameters) with two input sequence lengths (16k and 96k nucleotides).

The phylogenetic tree of the sampled species and the corresponding two-dimensional UMAP projections of sequence embeddings are shown in Figure~\ref{fig:model_overview}D. Across all four configurations, the embeddings form clear and well-separated clusters corresponding to the six taxonomic groups, indicating that the pre-trained model captures biologically meaningful, high-level features directly from raw genomic sequences. Notably, a pronounced scaling effect is observed: increasing model size from 1B to 3B parameters and extending input length from 16k to 96k nucleotides both lead to improved separation between taxonomic clusters.

Importantly, this structured clustering does not rely on local sequence homology. Genomic segments were sampled from random positions across the genomes of randomly selected species within each taxonomic group; they are neither aligned nor expected to share orthologous regions. The model’s ability to group such genomically disparate sequences by taxonomic origin suggests that it captures deep, genome-wide signatures of evolutionary relatedness that transcend specific loci or local sequence similarity. This observation highlights the capacity of next-token prediction pre-training to induce a generalized understanding of genomic organization and phylogenetic structure.

\begin{figure}[!htb]
    \centering
    \includegraphics[width=0.8\textwidth]{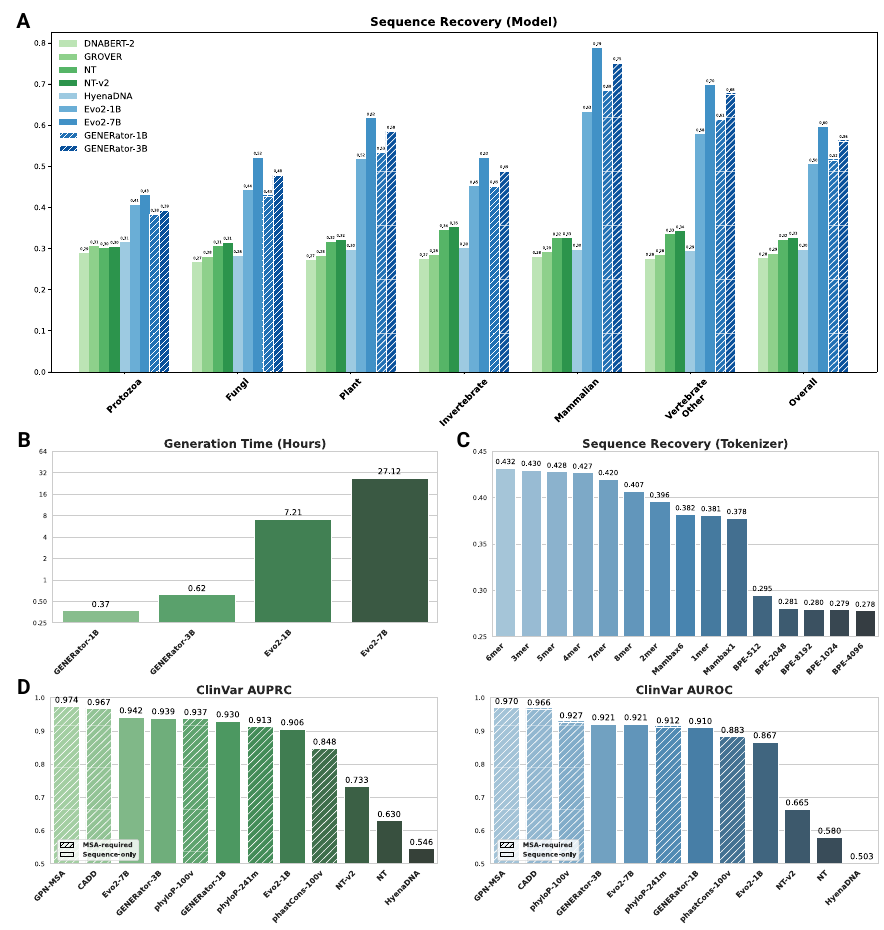}
    \caption{Training-free evaluations of GENERator.
    (A) Sequence recovery accuracy is compared across representative DNA foundation models. GENERator and Evo2 substantially outperform all other baseline models. At matched parameter scales, GENERator-1B consistently exceeds Evo2-1B in overall performance while achieving markedly higher generation efficiency. Scaling up to GENERator-3B yields clear performance gains over the 1B model and approaches the performance of Evo2-7B.
    (B) Generation time is measured on a single L40S GPU by conditioning on a 6k input sequence and generating the next 50 bp. GENERator exhibits tens-fold faster generation speed compared to Evo2 across model sizes, highlighting its substantially improved computational efficiency.
    (C) Sequence recovery performance is evaluated under identical model architectures, batch sizes, and training steps across six taxonomic groups. The 6-mer tokenizer achieves the best overall performance, whereas BPE tokenization performs consistently worse, likely due to the hierarchical nature of the BPE vocabulary.
    (D) Performance on ClinVar variant effect prediction is evaluated using AUPRC and AUROC. GENERator and Evo2 significantly outperform other sequence-based self-supervised models. GENERator-1B consistently surpasses Evo2-1B, while GENERator-3B achieves performance comparable to Evo2-7B. Although MSA-based methods such as GPN-MSA and CADD retain advantages on this task, GENERator and Evo2 operate without MSA information, enabling straightforward application to non-model organisms.}
    \label{fig:zeroshot_tasks}
\end{figure}

\subsection{Sequence Recovery}
\label{sec:kmer_predition}
\subsubsection{Tokenizer Comparison}
To identify an effective tokenization strategy for causal genomic language models, we conducted a controlled comparison of multiple tokenizers. All models share the same architecture as GENERator and are trained on identical datasets, differing only in the tokenization scheme, with evaluation performed after 32,000 training steps. Sequence recovery accuracy is used as the primary evaluation metric (Section~\ref{sec:sequence_recovery}). This task assesses a model’s ability to generate subsequent nucleotides within functional genomic regions given a prompt sequence, with performance quantified by overlap between generated and reference sequences. As a training-free evaluation, this setting enables an equitable comparison of intrinsic generative quality across different tokenization strategies.

A central consideration in tokenizer comparison is the trade-off between sequence resolution and contextual coverage. For transformer architectures, where computational cost scales quadratically with token count, comparisons are most appropriately conducted at a fixed number of input tokens rather than a fixed number of nucleotides. Under a fixed token budget (e.g., 16k tokens), a 6-mer tokenizer processes up to 96k nucleotides, whereas a single-nucleotide tokenizer is limited to 16k nucleotides. Increasing the 1-mer token count to match the nucleotide coverage of 6-mer tokenization would incur a 36-fold increase in computational cost, rendering such comparisons impractical due to both runtime and memory constraints. Accordingly, token-level evaluation provides the most realistic and operationally meaningful comparison framework.

Under this evaluation framework, the performance of $k$-mer tokenizers ($k=1$ to $8$) and BPE tokenizers (vocabulary sizes 512 to 8192) is summarized in Figure~\ref{fig:zeroshot_tasks}C, which reports the average sequence recovery accuracy across six taxonomic groups for a fixed input length of 1024 tokens and a prediction length of 30 nucleotides. The results reveal a clear performance optimum: neither the highest-resolution 1-mer tokenizer nor the longest-context 8-mer tokenizer achieves peak accuracy. Instead, the 6-mer tokenizer yields the strongest performance. We attribute this behavior to the fact that 6-mer tokenization strikes an effective balance between preserving local sequence grammar and enabling sufficient long-range contextual coverage under a fixed token budget. In contrast, smaller $k$ values limit the accessible context length, whereas larger $k$ values overly coarse-grain the sequence and obscure fine-grained regulatory patterns. This observation supports our systematic tokenizer selection strategy, in which empirical performance—guided by principled trade-offs—rather than architectural convention determines the final design choice.

BPE tokenizers, by comparison, consistently underperform across all tested vocabulary sizes, with sequence recovery accuracies below 0.30—only marginally above the random guessing baseline of 0.25. Although BPE increases nominal context length by grouping multiple nucleotides into single tokens, its hierarchical vocabulary interacts poorly with the next-token prediction objective. For instance, when the target next token corresponds to the nucleotide sequence \texttt{GCCT}, predictions such as \texttt{G}, \texttt{GC}, or \texttt{GCC} are deemed incorrect under the autoregressive objective, despite representing valid prefixes of the target sequence. This mismatch between hierarchical tokenization and the next-token prediction objective complicates optimization and likely contributes to the systematically inferior performance observed for BPE-based models.

It is worth noting that previous studies such as DNABERT-2~\cite{DNABERT-2} and GROVER~\cite{GROVER} systematically investigated tokenization strategies in the context of masked language modeling and both reported advantages for BPE. Taken together, our findings do not contradict these prior results, but instead highlight that the effectiveness of a given tokenization scheme is strongly conditioned on the underlying training objective. 

In particular, the degradation observed for BPE in our experiments is specific to autoregressive (next-token prediction) training. Under a masked language modeling objective, each masked position is associated with exactly one correct target token, eliminating ambiguity arising from hierarchical tokenization. This distinction explains why BPE performs well for masked genomic language models, yet fails to generalize to causal sequence generation, where token-level ambiguity directly interferes with next-token prediction.

More broadly, these results underscore a key methodological point: conclusions drawn from large language models trained on natural language, or from MLM-based genomic language models, do not necessarily transfer to causal genomic language models. Instead, optimal tokenization strategies depend critically on the choice of training objective, emphasizing the importance of evaluating tokenization schemes in task-aligned settings.

\subsubsection{Architecture Comparison}
To achieve long-context modeling in genomics, there are two natural strategies: (i) retain the transformer architecture and increase effective context coverage via coarser tokenization (e.g., $k$-mers), trading off single-nucleotide input resolution; or (ii) retain single-nucleotide tokenization and adopt a more computationally efficient architecture that can process substantially longer token sequences. State space models (SSMs), such as Mamba-2~\cite{Mamba,Mamba-2}, are a prominent candidate for the latter strategy. We therefore evaluated whether an SSM can convert longer single-nucleotide context into improved generative fidelity on the sequence recovery task.

We trained a 1.2B-parameter Mamba-2 model with single-nucleotide tokenization and a nominal context length of 98k nucleotides, and defined two comparison schemes:
\begin{itemize}
    \item \textbf{Mamba$\times$1}: Processes the same number of input tokens as the transformer-based models, matching the nucleotide coverage of a transformer with 1-mer tokenization.
    \item \textbf{Mamba$\times$6}: Processes six times more input tokens than the transformer-based models, matching the nucleotide coverage of a transformer with 6-mer tokenization.
\end{itemize}

The results are shown in the right panel of Figure~\ref{fig:zeroshot_tasks}C. Under the first scheme, Mamba$\times$1 achieved a sequence recovery accuracy of 0.378, slightly below the 0.381 achieved by the transformer with 1-mer tokenization at the same input length. Under the second scheme, Mamba$\times$6 reached an accuracy of 0.382—only a marginal improvement of 0.001 over the transformer with 1-mer tokenization—despite a six-fold increase in nominal context length. When compared to the transformer with 6-mer tokenization at equivalent nucleotide input (96k nucleotides), Mamba$\times$6 showed substantially lower performance (0.382 vs.\ 0.432).

Together, these results indicate that increasing single-nucleotide context via an SSM does not yield the expected gains in sequence recovery accuracy. In Mamba-2, and more broadly in this class of SSM architectures, the entire input sequence—regardless of its length—is recursively compressed into a fixed-dimensional hidden state representation. As the sequence length increases, this fixed-dimensional representation saturates, limiting the model’s ability to retain and exploit long-range dependencies~\cite{Empirical, DeciMamba}. As a result, extending nominal context length alone does not translate into improved generative fidelity. For practical long-context genomic generation in this study, we therefore retain the transformer-based approach with $k$-mer tokenization for subsequent experiments.

Comprehensive results, including analyses across varying input token lengths, prediction lengths, and detailed breakdowns for each taxonomic group, are provided in Figure~\ref{fig:kmer_tokenizer}.

\subsubsection{Baseline Model Comparison}
To comprehensively benchmark GENERator against existing genomic foundation models, we designed a standardized evaluation protocol with fixed input and prediction lengths across all models. Specifically, each model was tasked with processing 6,144 nucleotides of prompt sequence and generating the subsequent 30 nucleotides. This setting reflects practical application scenarios in which sequence lengths are predetermined and users may switch between models with different tokenization schemes. Under this protocol, GENERator with 6-mer tokenization processes 1,024 input tokens, whereas models such as Evo2 with single-nucleotide tokenization process the full 6,144 tokens. This design enables a realistic comparison of generative accuracy while allowing direct assessment of computational efficiency under identical sequence-level conditions.

To ensure a phylogenetically balanced evaluation, we uniformly sampled 5,000 sequences from each of six eukaryotic taxonomic groups, yielding a total of 30,000 test sequences. Sequence recovery accuracy for each taxonomic group, as well as the overall average across all groups, is shown in Figure~\ref{fig:zeroshot_tasks}A. A more comprehensive analysis across varying input lengths and prediction lengths is provided in the grid plot in Figure~\ref{fig:kmer_model}.

In terms of generative accuracy, GENERator exhibits strong and consistent performance across all taxonomic groups. As summarized by the overall results in Figure~\ref{fig:zeroshot_tasks}A, GENERator-1B achieves an accuracy of 0.52, outperforming Evo2-1B (0.50). Scaling to GENERator-3B further improves accuracy to 0.56, approaching the performance of Evo2-7B (0.60) despite a substantially smaller parameter count.

The advantage of GENERator becomes even more pronounced in computational efficiency. The total wall-clock time required to process all 30,000 sequences is shown in Figure~\ref{fig:zeroshot_tasks}B. GENERator-1B completed the evaluation in 0.37 hours, and GENERator-3B in 0.62 hours. In contrast, Evo2-1B required 7.21 hours—over 19$\times$ slower than GENERator-1B while delivering lower accuracy. Evo2-7B required 27.12 hours, corresponding to a 44$\times$ slowdown relative to GENERator-3B for only a modest gain in accuracy. The Evo2-40B model could not be evaluated locally due to prohibitive computational requirements; attempts using the NVIDIA Evo2-40B API (\url{https://build.nvidia.com/arc/evo2-40b}) consistently encountered a 120-second timeout limit per request, highlighting the practical scalability challenges associated with extremely large genomic models.

Notably, GENERator and Evo2 form a distinct top performance tier, substantially outperforming all other baseline models (Figure~\ref{fig:zeroshot_tasks}A). Among masked language models (MLMs; shown in green), the Nucleotide Transformer (NT) series achieves improved performance relative to other MLMs but remains markedly below the leading causal language models. This gap is consistent with the limitations of masked modeling objectives for open-ended sequence generation. HyenaDNA, despite adopting an autoregressive objective, does not outperform MLM-based models, a result likely attributable to its comparatively small model size (55M parameters).

\subsection{Variant Effect Prediction}
Variant effect prediction (VEP) aims to assess whether a single-nucleotide variant (SNV) is benign or deleterious, and is therefore intrinsically a single-base resolution task. At first glance, such tasks appear poorly matched to models trained with $k$-mer tokenization, which trades input resolution for increased contextual coverage. To address this mismatch, we compute marginal nucleotide-level probabilities by aggregating token-level predictions, thereby recovering single-nucleotide resolution while retaining the benefits of long-context modeling. The detailed formulation of this marginalization strategy is described in Section~\ref{sec:variant-effect-prediction}; here, we focus on its empirical performance.

Building upon this framework, we evaluated the VEP capabilities of GENERator and comparative models using the ClinVar~\cite{clinvar} dataset curated in GPN-MSA~\cite{gpn-msa}. Where applicable, baseline results are taken directly from the GPN-MSA paper.

To ensure a fair and architecture-appropriate evaluation, we adapted the inference strategy for each model class. For masked language models, including the NT series, we leveraged bidirectional attention by centering the variant site within the input sequence, masking the target nucleotide, and computing probabilities for all possible alleles. For autoregressive models such as GENERator and Evo2, we employed causal attention by positioning the variant site at the sequence end and performing next-token prediction. For models utilizing $k$-mer tokenization (including the NT and GENERator series), we applied the token probability marginalization approach described in Section~\ref{sec:variant-effect-prediction}, which improves performance over naive token-level probability comparison. Accordingly, we updated the NT series results reported in the original GPN-MSA study.

Results are summarized in Figure~\ref{fig:zeroshot_tasks}D. At comparable parameter scales, GENERator-1B outperforms Evo2-1B, while GENERator-3B achieves performance competitive with Evo2-7B, highlighting the effectiveness of our approach. Other sequence-based self-supervised models, including the NT series and HyenaDNA, exhibit substantially lower performance.

As expected, MSA-based methods such as GPN-MSA and CADD~\cite{cadd} maintain a performance advantage on this task, benefiting from explicit evolutionary information encoded in multiple sequence alignments. However, these approaches are limited by the availability of high-quality alignments and are therefore primarily applicable to well-studied organisms such as humans, mammals, and other vertebrates. In contrast, traditional conservation-based MSA methods such as phyloP~\cite{phylop} and phastCons~\cite{phastcons} are matched or surpassed by both GENERator and Evo2, indicating that modern genomic language models can capture meaningful evolutionary constraints from single-sequence information alone.

The alignment-free nature of GENERator and Evo2 enables principled application across a wide range of species, including insects, plants, and fungi, where MSA-based methods are often infeasible due to the lack of comparative genomic data. Although comprehensive validation beyond human datasets remains challenging because of the scarcity of annotated variant resources such as ClinVar, this capability represents a significant practical advantage for variant effect prediction in non-model organisms. Collectively, these results position GENERator as a scalable and effective alignment-free alternative for VEP, particularly in settings where MSA-based approaches are impractical or unavailable.

\begin{figure}[!htb]
    \centering
    \includegraphics[width=0.8\textwidth]{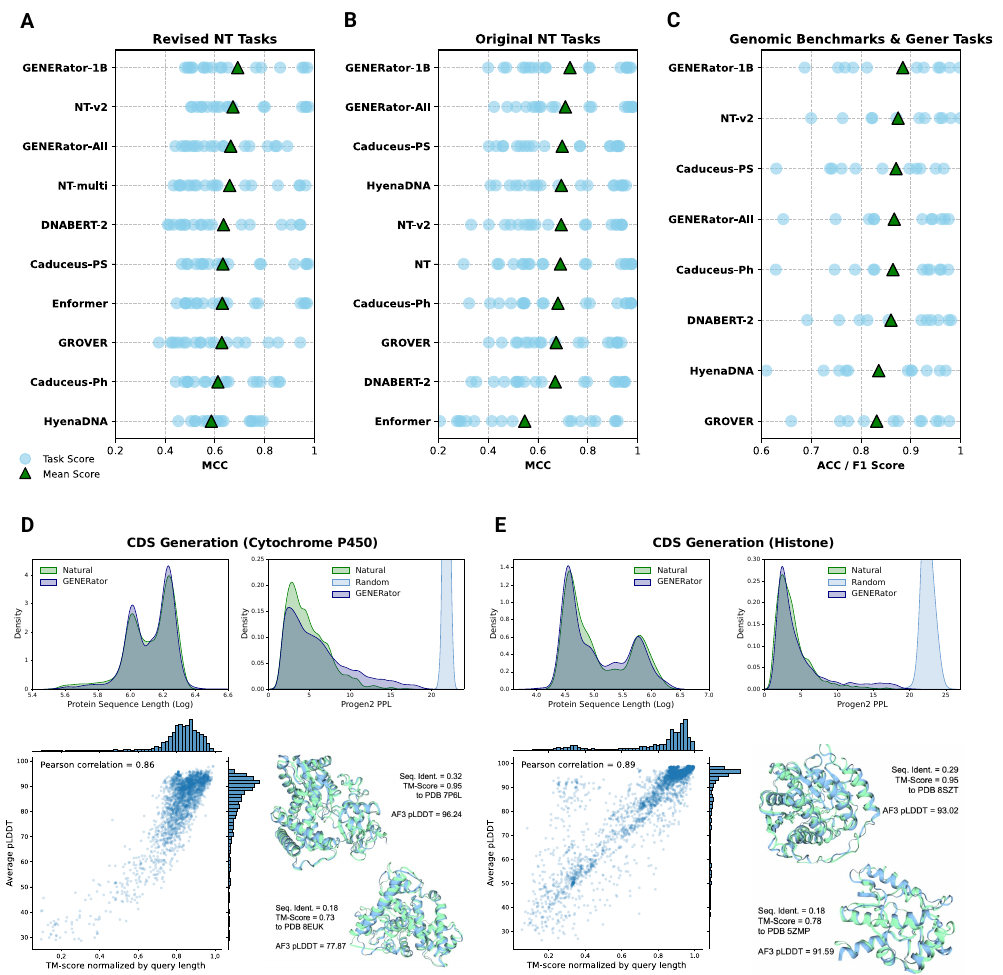}
    \caption{Task-specific fine-tuning of GENERator.
    (A-C) Comparison across diverse supervised benchmarks, including Revised NT tasks (A), Original NT tasks (B), Genomic Benchmarks and Gene tasks (C), shows that GENERator achieves the strongest overall performance across all evaluated settings.
    (D-E) Central-dogma tasks are evaluated on the cytochrome P450 family (D) and histone family (E). Generated DNA sequences exhibit stable and continuous coding structures. Translated proteins show low perplexity under the protein language model ProGen2, indicating high protein-likeness. Structural validation using AlphaFold3 followed by Foldseek reveals that most generated proteins fall within high-confidence regions. Importantly, many generated proteins display strong structural similarity to natural counterparts despite low sequence identity ($<0.3$), demonstrating that GENERator captures protein grammar rather than memorizing natural sequences.}
    \label{fig:finetune_tasks}
\end{figure}

\subsection{Benchmark Evaluations}
\label{sec:benchmark_exp}
In addition to training-free evaluations of intrinsic model capabilities, task-specific fine-tuning is a widely used paradigm for deploying foundation models across diverse downstream tasks. To assess the adaptability of GENERator in this regime, we conducted a comprehensive evaluation on a suite of established genomic benchmarks (Section~\ref{sec:benchmark}). This evaluation serves as a critical bridge, validating the model’s performance on standardized supervised tasks before its application to more specialized real-world problems, such as protein and enhancer design (Sections~\ref{sec:central_dogma_exp} and~\ref{sec:enhancer_design_exp}).

We compared GENERator with a range of representative genomic foundation models, including Enformer~\cite{enformer}, DNABERT-2, HyenaDNA, Nucleotide Transformer (NT), Caduceus, and GROVER. To ensure a fair comparison, all models were fine-tuned and evaluated using 10-fold cross-validation on each dataset. For every model–dataset pair, we performed extensive hyperparameter searches over learning rates $\{1\times10^{-5}, 2\times10^{-5}, 5\times10^{-5}, \ldots, 1\times10^{-3}, 2\times10^{-3}, 5\times10^{-3}\}$ and batch sizes $\{64, 128, 256, 512\}$. Detailed configurations and implementation details are provided in Section~\ref{sec:exp_config_supp}.

\paragraph{NT Tasks.}
Given the recent revision of the NT task dataset~\cite{nucleotide-transformer}, we evaluated models on both the original and revised benchmark sets. The average performance across tasks in each benchmark is shown in Figure~\ref{fig:finetune_tasks}A and B, with per-task results reported in Tables~\ref{tab:nucleotide_transformer_tasks_revised} and~\ref{tab:nucleotide_transformer_tasks}. Across both versions, GENERator exhibits the strongest average performance among evaluated models. Notably, the reduced performance observed for GENERator-All highlights the importance of the functional sequence training strategy (Section~\ref{sec:data_preparation}). Enformer maintains competitive performance on chromatin profile and regulatory element tasks, reflecting its original supervised training focus on these domains~\cite{enformer}. The recently released Nucleotide Transformer v2 (NT-v2), despite its smaller size (500M parameters), achieves higher average performance than the original NT model (2.5B parameters). In contrast, models employing BPE tokenization (DNABERT-2, GROVER) or single-nucleotide tokenization (HyenaDNA, Caduceus) do not show consistent advantages on these benchmarks.

\paragraph{Genomic Benchmarks \& Gener Tasks.}
We further evaluated models on Genomic Benchmarks~\cite{genomic-benchmarks}, which primarily target human genomic tasks, and on the newly proposed Gener tasks, designed to assess gene type and taxonomic classification across diverse organisms and sequence lengths. Average performance across all tasks in both benchmark suites is summarized in the right panel of Figure~\ref{fig:finetune_tasks}C, with detailed results reported in Tables~\ref{tab:genomic_benchmarks} and~\ref{tab:gener_tasks}.

On Genomic Benchmarks, GENERator achieves the highest overall performance. Caduceus models perform competitively despite their small size (8M parameters), likely due to their specialization on human genomic data; however, this specialization appears to limit generalization, as reflected by weaker performance on multi-species tasks. On the Gener tasks, GENERator attains the strongest results on both gene classification and taxonomic classification, with NT-v2 also performing well. These results are consistent with the benefits of comprehensive multi-species pre-training. DNABERT-2 shows reduced performance, potentially attributable to its limited model size (117M parameters) and short context length (3k nucleotides). Models trained primarily on human genomes (HyenaDNA, Caduceus) still exhibit some degree of generalization after fine-tuning, whereas GROVER exhibits markedly reduced performance on taxonomic classification, likely due to its restricted context length (3k nucleotides).

\paragraph{Note on Model Selection and Scale.}
The recently released Evo2 series~\cite{Evo2} is not included in these fine-tuning evaluations for two practical reasons. First, open-source implementations for fine-tuning Evo2 are currently unavailable. Second, the computational requirements pose a prohibitive barrier: as shown in our sequence recovery benchmarks (Figure~\ref{fig:zeroshot_tasks}B), Evo2 models require tens-fold more inference time than GENERator. This gap would be further amplified during fine-tuning, rendering comprehensive evaluation impractical for most research settings, including ours.

For similar considerations, fine-tuning evaluations in this study are conducted using the GENERator-1B model. Although scaling trends observed in zero-shot tasks suggest that the GENERator-3B model would achieve higher performance, the associated increase in computational cost was deemed disproportionate to the expected gains for the purpose of comparative benchmarking.

\subsection{Central Dogma}
\label{sec:central_dogma_exp}
Beyond conventional benchmarking tasks, we designed the Central Dogma experiment to assess whether GENERator can capture fundamental biological principles through DNA sequence modeling alone (Section~\ref{sec:central_dogma}). Specifically, this task evaluates whether a generative genomic language model trained exclusively on DNA sequences can generate protein-coding DNA that translates into structurally plausible proteins, consistent with the flow of genetic information from DNA to protein.

We selected two evolutionarily distinct protein families, cytochrome P450 and histone, and curated their protein-coding DNA sequences from RefSeq via UniProt mappings. After fine-tuning GENERator on these family-specific datasets, we generated novel DNA sequences and evaluated the quality of the resulting constructs using multiple complementary criteria.

As shown in Figure~\ref{fig:finetune_tasks}D for cytochrome P450 and Figure~\ref{fig:finetune_tasks}E for histone, the length distributions of both generated DNA sequences and their translated protein products closely match those of natural sequences after deduplication. Critically, 99.9\% of generated sequences were successfully translated into proteins using standard codon tables, indicating stable coding structure without premature stop codons or frameshift errors. This high translation success rate demonstrates that the model preserves essential constraints of protein-coding regions.

We further performed protein-level validation to assess the structural and statistical plausibility of the generated sequences. Sequence naturalness, evaluated using the protein language model ProGen2~\cite{progen2}, showed that perplexity distributions of generated proteins closely align with those of natural family members, while remaining clearly distinct from shuffled controls. Structural modeling with AlphaFold3 revealed that the generated sequences fold into stable tertiary structures, and FoldSeek~\cite{Foldseek} analysis identified high structural similarity to known Protein Data Bank folds ($\text{TM-score} > 0.8$), despite low sequence identity ($< 0.3$). This combination of structural conservation and sequence divergence indicates that GENERator captures general protein-coding and folding principles rather than memorizing existing sequences.

Together, these results demonstrate that GENERator can generate protein-coding DNA sequences that translate into structurally plausible proteins, supporting the feasibility of DNA-level generative modeling for protein design. While protein language models such as ProGen2 enable direct generation at the amino acid level, DNA-level generation offers a complementary representation that naturally accommodates codon-level structure. Owing to the degeneracy of the genetic code, identical protein sequences can be encoded by multiple DNA sequences with different codon usage patterns, which vary across organisms. By operating directly in DNA sequence space, generative models such as GENERator have the potential to capture and exploit such codon-level regularities. Exploring explicit codon optimization or organism-specific design strategies within this framework remains an important direction for future work.

\begin{figure}[!htb]
    \centering
    \includegraphics[width=0.8\textwidth]{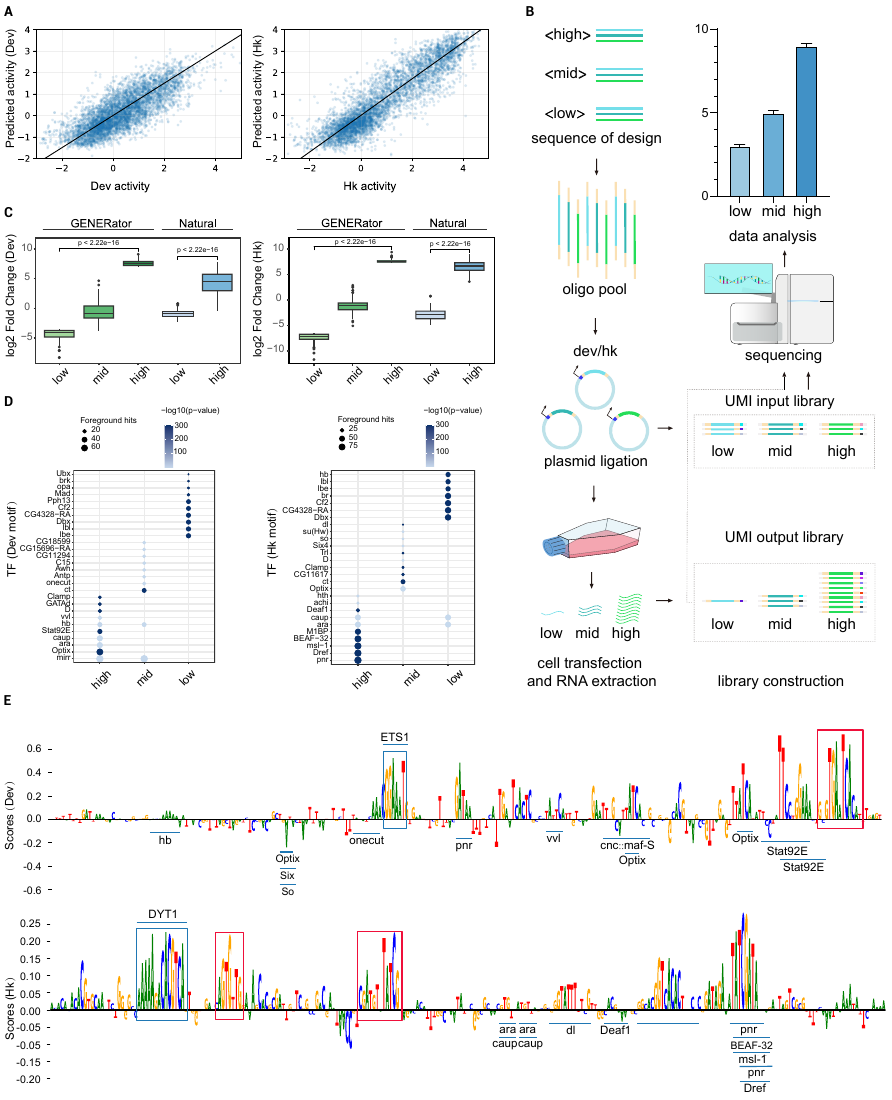}
    \caption{GENERator enables accurate design of CREs with dynamic ranges exceeding those of natural sequences. (A) Correlation between experimentally measured and model-predicted activities of CREs in the DeepSTARR hold-out test set. (B) Schematic overview of the UMI-STARR-seq library construction workflow using developmental (DSCP; Dev) and housekeeping (RpS12; Hk) core promoters. GENERator-designed sequences were synthesized as pooled oligonucleotides, cloned into reporter plasmids, and transfected into S2 cells. CRE activity was quantified as enrichment of UMI counts in RNA output relative to plasmid input. (C) Comparison of experimental activities between GENERator-designed and natural sequences. For GENERator-designed sequences: top 100 from the \texttt{<high>} group, bottom 100 from the \texttt{<low>} group, and 100 randomly selected from the \texttt{<mid>} group. For natural sequences: top 100 highest-activity and bottom 100 lowest-activity sequences from the DeepSTARR dataset. (D) Transcription factor DNA-binding motifs enriched in GENERator-designed sequences of different activity levels. (E) Motif composition of two representative GENERator-designed sequences. Blue lines: insect transcription factor motifs; blue boxes: vertebrate transcription factor motifs; red boxes: unannotated high-contribution regions.}
    \label{fig:enhancer_design}
\end{figure}

\subsection{Cis-Regulatory Element Design with High-throughput Validation}
\label{sec:enhancer_design_exp}
Functional DNA design represents a compelling application domain for generative genomic language models, with the potential to move beyond descriptive modeling toward direct sequence engineering. A notable example is the design of CRISPR-Cas systems by Evo~\cite{Evo}, which demonstrated that language models trained on genomic sequences can be used to generate functional DNA with experimentally validated activity. In the Evo study, the model was fine-tuned on CRISPR-Cas loci and used to generate sequences in an unconditional manner, producing over two million candidate sequences. Through multiple rounds of computational and experimental screening, 11 candidates were selected for wet-lab validation, among which EvoCas9-1 achieved cleavage efficiency comparable to SpCas9 systems~\cite{hart2017evaluation}. 

From a methodological perspective, however, the Evo workflow relies on largely unguided sampling, with functional bias introduced primarily through extensive post hoc screening rather than during the generation process. This paradigm is analogous in spirit to traditional template-based random mutagenesis coupled with functional selection. While effective, such unconditional generation offers limited leverage for targeted design or optimization.

These considerations motivate the development of generative frameworks that support explicit, preference-guided sequence design. In this work, we explore such a framework in the context of cis-regulatory element (CRE) design. Unlike protein-coding sequences with well-defined amino acid mappings, non-coding regulatory elements—including enhancers and silencers—operate through complex cis-regulatory grammars that lack straightforward sequence-to-function relationships. This inherent complexity makes CREs a particularly suitable testbed for conditional generative design, where generation-time bias is essential for efficient exploration of functional sequence space.

Our CRE design framework consists of two complementary components: an activity-conditioned CRE generator that enables prompt-guided sequence design, and a CRE activity predictor that provides fine-grained quantitative scoring for candidate selection. Both components are derived by fine-tuning GENERator on the DeepSTARR dataset~\cite{DeepSTARR}, which provides large-scale quantitative measurements of CRE activity.

Following the sequence design protocol described in Section~\ref{sec:sequence_design}, we first fine-tune GENERator as an activity-conditioned sequence generator by associating discrete activity labels with prefix tokens (\texttt{<high>}, \texttt{<mid>}, \texttt{<low>}). This generator implements coarse-grained control over CRE activity at generation time, enabling prompt-guided design of sequences biased toward high, medium, or low regulatory activity. Using this generator, we produced approximately 40{,}000 unique CRE sequences for both developmental (Dev) and housekeeping (Hk) promoters under the three activity conditions.

To enable fine-grained ranking within each activity regime, we additionally fine-tune GENERator as a supervised CRE activity predictor that maps input sequences to continuous activity scores. On the held-out DeepSTARR test set, this predictor achieved Pearson correlation coefficients of 0.71 for developmental CREs and 0.80 for housekeeping CREs (Figure~\ref{fig:enhancer_design}A), representing a substantial improvement over previously evaluated models (Table~\ref{tab:enhancer_benchmark}).

Candidate sequences generated under each prefix condition were then ranked using the CRE activity predictor. For experimental validation, we selected the top 1{,}900 sequences from the \texttt{<high>} group, the bottom 1{,}900 sequences from the \texttt{<low>} group, and 1{,}900 randomly sampled sequences from the \texttt{<mid>} group. In addition, we included 400 natural genomic sequences (the top and bottom 100 from each promoter class) from the DeepSTARR dataset, as well as 26 sequences designed by the DREAM model~\cite{li2024novel}. In total, 12{,}026 oligonucleotides were synthesized and evaluated using UMI-STARR-seq assays under both the Dev (DSCP) and Hk (Rps12) promoters in \emph{Drosophila melanogaster} S2 cells (Figure~\ref{fig:enhancer_design}B).

Consistent with model predictions, experimentally measured activities showed a strong correlation with predicted scores, reaching $R^2 = 0.48$ for Dev CREs and $R^2 = 0.83$ for Hk CREs (Figure~\ref{fig:enhancer_supp_2}B). Moreover, GENERator-designed sequences generated with different prefix tokens (\texttt{<high>}, \texttt{<mid>}, and \texttt{<low>}) displayed clearly separated activity distributions in the UMI-STARR-seq assays. 

Notably, both the highest- and lowest-activity GENERator-designed sequences extended beyond the activity range observed for natural genomic sequences (Figure~\ref{fig:enhancer_design}C). To ensure a meaningful comparison, we selected the top 100 and bottom 100 natural sequences from the DeepSTARR dataset of over 400,000 sequences (labeled Natural-high and Natural-low), and compared them with the top 100 and bottom 100 sequences from the GENERator-designed \texttt{<high>} and \texttt{<low>} groups, respectively, along with 100 randomly sampled sequences from the \texttt{<mid>} group. It is important to note that while natural sequences were selected from a pool of over 400,000, our designed sequences were selected from only approximately 11,600 experimentally tested sequences. Despite this substantial difference in selection pool sizes, GENERator-designed sequences still exhibited significantly broader activity ranges than their natural counterparts, underscoring the exceptional efficiency and potency of our design approach.

For Hk CREs, the highest-activity sequence designed by GENERator exceeded the strongest natural genomic sequence by 35\% and outperformed the best sequence designed by DREAM~\cite{li2024novel} by 78\% (Figure~\ref{fig:enhancer_supp_3}A). For Dev CREs, the top GENERator-designed enhancer achieved more than a two-fold increase in activity relative to the strongest natural sequence, although its activity remained 45\% lower than that of the top DREAM-designed sequence.

Beyond the design of highly active enhancers, GENERator also demonstrated the ability to perform inverse design by generating strong silencer-like CREs. Specifically, the lowest-activity sequences generated under the \texttt{<low>} prompt exhibited activities more than 100-fold lower than the weakest natural sequences in the DeepSTARR dataset. Together, these results indicate that GENERator can robustly explore both extremes of the cis-regulatory activity spectrum, spanning strong enhancers and ultra-low-activity silencers.

Collectively, these findings demonstrate that GENERator can explore regulatory sequence space beyond naturally occurring cis-regulatory elements. The identification of both high-activity enhancers and low-activity silencers whose regulatory strengths extend beyond the natural genomic range highlights the potential of generative genomic language models to not only reproduce, but also expand the functional repertoire of regulatory genomics.

Motif enrichment analysis revealed both shared and distinct enrichment patterns of transcription factor DNA-binding motifs in high-activity Dev and Hk sequences. For instance, \emph{ara} and \emph{caup} are enriched in both high-activity Dev and Hk sequences. In contrast, transcription factors such as \emph{mirr}, \emph{Optix}, and \emph{Stat92E} are specifically enriched in high-activity Dev sequences, while \emph{pnr}, \emph{Dref}, and \emph{msl-1} are enriched in high-activity Hk sequences. Similar patterns of shared and specific motif enrichment were observed in low-activity Dev and Hk sequences. These findings suggest that CRE--promoter compatibility varies between Hk and Dev promoters, and that GENERator successfully captures this biological phenomenon during training (Figures~\ref{fig:enhancer_design}D, \ref{fig:enhancer_supp_3}B).

Using base-level contribution scores computed as described in Section~\ref{sec:sequence_design}, we identified nucleotide positions that are critical for CRE activity (Figure~\ref{fig:enhancer_design}E). Applying this approach to both a strong Dev enhancer and a strong Hk enhancer revealed pronounced contributions from motifs corresponding to known transcriptional activators. Specifically, scanning CRE sequences with insect transcription factor DNA-binding motifs identified several well-characterized \emph{D. melanogaster} transcription factors, including \emph{Optix}, \emph{hb}, \emph{Stat92E}, \emph{M1BP}, \emph{pnr}, \emph{Dref}, \emph{ara}, and \emph{caup}~\cite{li2013optix, staller2015shadow, tsurumi2011stat, poliacikova2023m1bp, bag2021m1bp, calleja2000generation, tsuchiya2007transcriptional, gomez1996araucan}.

Given that insect transcription factor motif databases are relatively less comprehensive than those for humans and other well-studied organisms, we additionally scanned vertebrate motif collections and identified motifs associated with other known transcriptional activators, such as \emph{ETS1} and \emph{DYT1}~\cite{chen2017vegf, cui2016feedback}. Notably, even after this systematic screening, we observed high-contribution sequence regions that did not correspond to any known transcription factor DNA-binding motifs, suggesting that GENERator captures regulatory sequence features beyond canonical TF motifs and motivating further mechanistic investigation (Figure~\ref{fig:enhancer_design}E).

The computational-experimental workflow established here represents a significant advancement over traditional approaches. Compared to random mutagenesis or unconditional generation methods, our prompt-guided framework enables targeted exploration of sequence space with specific functional objectives. This `educated guess' approach dramatically improves sampling efficiency, with the high-activity group showing 15-fold higher enrichment over random sampling. Collectively, these results demonstrate that GENERator enables the efficient and precise design of CREs with diverse regulatory activities that surpass those of natural genomic sequences, highlighting its substantial potential for applications in synthetic biology and gene therapy. In addition, GENERator offers a powerful framework for uncovering novel principles underlying transcriptional regulatory mechanisms.

\section{Discussion}
In this study, we introduced GENERator, a generative genomic foundation model trained on 386 billion nucleotides of eukaryotic DNA, and systematically evaluated its capabilities across representation learning, sequence generation, and experimentally validated regulatory design. Beyond reporting improved performance, this work uses GENERator as a lens to examine how specific modeling choices—spanning data curation, tokenization, architecture, and inference—determine whether large-scale sequence models can capture biologically meaningful structure and support practical genomic applications.

\subsection{Key Contributions}
The contributions of this work can be summarized along several interrelated dimensions, each highlighting a distinct aspect of effective generative genomic modeling.

\paragraph{First, biologically informed training data is critical for genomic foundation models.}
We introduce a functional sequence training paradigm that emphasizes gene-centric functional regions while avoiding overwhelming the model with non-functional repetitive content. Models trained under this paradigm consistently learn more effective representations and generalize more robustly across downstream tasks than those trained on indiscriminate genomic sequences. This result underscores that, in genomics, scaling data volume alone is insufficient; aligning training data with biological function plays a decisive role in shaping model behavior.

\paragraph{Second, tokenization is a domain-specific design choice that fundamentally shapes generative performance.}
While byte pair encoding (BPE) is advantageous in natural language processing, we find that it performs poorly in autoregressive DNA sequence generation. The hierarchical and overlapping structure of BPE vocabularies introduces ambiguity at the nucleotide level during next-token prediction, a mismatch that is largely absent in human language but becomes detrimental in genomic contexts. In contrast, $k$-mer tokenization provides a more suitable inductive bias, balancing local sequence resolution with contextual coverage. This finding highlights a fundamental distinction between linguistic and biological sequence statistics and demonstrates that tokenization strategies do not necessarily transfer directly across domains.

\paragraph{Third, effective long-context modeling depends on how context is represented and integrated, not on nominal context length alone.}
We explicitly examined an alternative modeling strategy that retains single-nucleotide resolution while adopting more computationally efficient state space models (SSMs) such as Mamba-2. Our results show that, in this setting, substantially increasing nominal single-nucleotide context does not translate into proportional gains in generative accuracy. Although SSMs can ingest longer sequences, our results indicate that the additional context is not effectively leveraged for sequence recovery under an autoregressive generative objective. In contrast, transformer-based models paired with appropriate tokenization more reliably convert extended context into biologically informative representations.

\paragraph{Fourth, DNA-level generative models can internalize higher-order biological constraints.}
Despite being trained exclusively on DNA sequences, GENERator generates protein-coding DNA that translates into structurally plausible proteins, indicating that aspects of the central dogma can emerge naturally from sequence-level objectives. In parallel, through probabilistic marginalization, GENERator supports alignment-free variant effect prediction at single-nucleotide resolution, demonstrating that reduced input-level resolution does not preclude precise base-level inference.

\paragraph{Finally, generative genomic models can be coupled with task-aligned prompting and experimental validation to enable controllable biological design.}
Using a prompt-guided framework, GENERator efficiently explores cis-regulatory sequence space, producing both highly active enhancers and ultra-low-activity silencer-like elements that are validated by high-throughput UMI-STARR-seq assays. These results illustrate how generative models can move beyond descriptive modeling toward targeted, experimentally grounded sequence design.

The generality of these insights is further supported by external validation on the Genomic Touchstone benchmark~\cite{genomic_touchstone}, where GENERator achieves top-tier performance across DNA-, RNA-, and protein-level tasks. Taken together, this work demonstrates that aligning model design with biological structure—rather than pursuing scale in isolation—is central to building genomic foundation models that are both scientifically grounded and practically useful.

\subsection{Limitations and Future Directions}
Despite the advances demonstrated by GENERator, several limitations define the current scope of this work. Most notably, the present model is trained exclusively on eukaryotic genomic sequences and does not incorporate prokaryotic or viral genomes. This focus reflects fundamental differences in genome organization, regulatory logic, and functional annotation across domains of life, and was a deliberate choice to ensure effective modeling within a biologically coherent setting.

Recent efforts such as Evo2~\cite{Evo2} adopt a unified modeling strategy that spans eukaryotic, prokaryotic, and viral genomes. While this approach provides broad biological coverage, it also incurs substantial computational cost in terms of model size, training resources, and inference efficiency, which can limit accessibility and make systematic experimentation challenging in practice. In this work, we demonstrate that strong generative performance can be achieved within a well-defined biological domain—specifically, eukaryotic genomes—without resorting to extreme model scale or prohibitive computational cost.

Building on this observation, a natural direction for extending GENERator beyond its current scope is to explore an evolutionary-informed modular framework. Such a framework would decompose genomic modeling into biologically meaningful components, with each module specializing in genomic contexts defined by shared evolutionary and functional characteristics.

\begin{figure}[ht]
    \centering
    \includegraphics[width=0.7\textwidth]{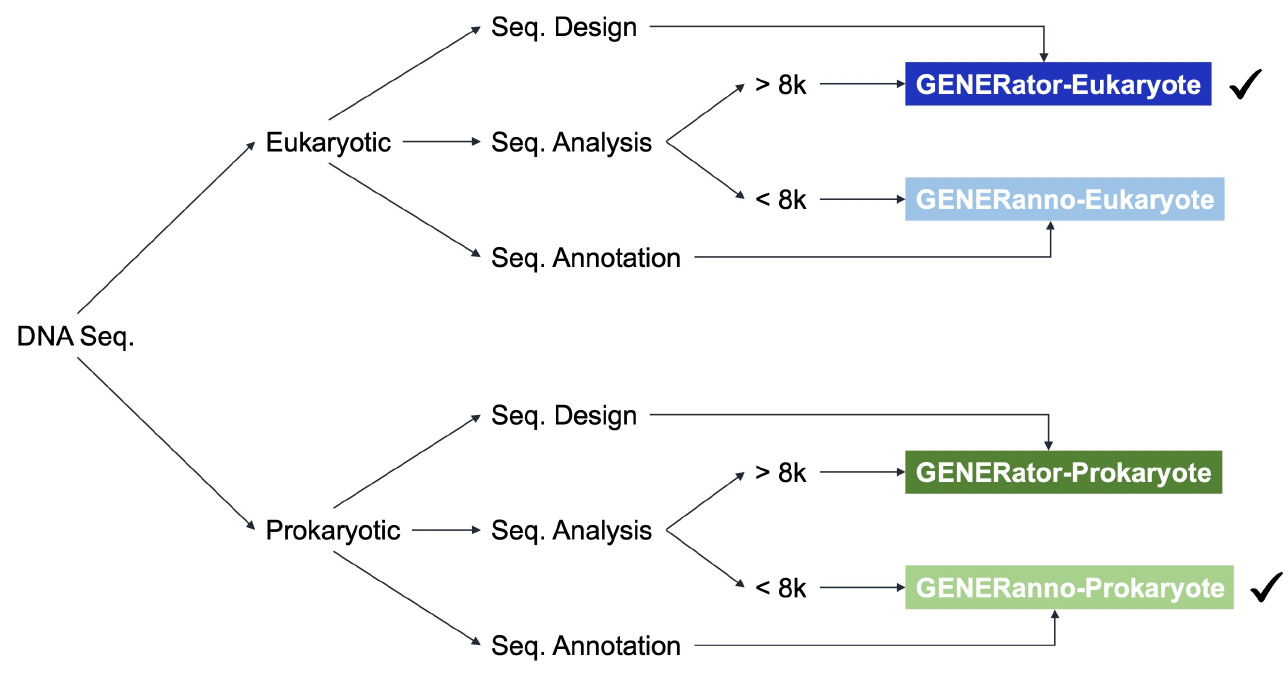}
    \caption{Overview of the Gener Project. This figure illustrates the modular architecture of the Gener Project, dividing tasks among four distinct experts. This design enables individual models to be deployed, updated, and scaled independently, simplifying maintenance and reducing resource demands.}
    \label{fig:gener_universe}
\end{figure}

As illustrated in Figure~\ref{fig:gener_universe}, our current ecosystem already reflects elements of this modular perspective, comprising a generative model optimized for long-context eukaryotic sequences (GENERator-Eukaryote) alongside a complementary module tailored to prokaryotic and metagenomic annotation tasks (GENERanno-Prokaryote~\cite{li2025generanno}). Future work may explore additional extensions, such as generative modeling for prokaryotic genomes or annotation-oriented models for specific classes of eukaryotic regulatory elements, enabling each component to be developed, scaled, and evaluated according to its biological and computational requirements.

Viral genomes present an additional modeling challenge due to their strong dependence on host cellular machinery~\cite{correa2021revisiting}. Rather than treating viruses as an isolated domain, a promising direction is to model viral sequences in conjunction with their host genomic context, for example through continued or conditional pre-training. Such approaches may better capture virus--host interactions and co-evolutionary dynamics, while remaining compatible with a modular modeling framework.

Taken together, these considerations outline how the framework established by GENERator may be extended beyond its current scope. Within the context of this work, evolutionary-informed modular modeling represents a practical and biologically grounded direction for scaling generative genomic models while balancing domain specificity and computational feasibility.

\subsection{Broader Implications}
The development of GENERator reflects a broader vision for practical and accessible AI for biology. In recent years, the field has increasingly pursued larger model scales, as exemplified by ESM3~\cite{esm3}, ProGen3~\cite{ProGen3}, xTrimoPGLM~\cite{xTrimoPGLM}, and Evo2~\cite{Evo2}, with parameter counts reaching tens or even hundreds of billions. While such models have achieved impressive results, their reliance on extreme scale introduces substantial barriers to deployment, particularly for research settings with limited computational resources.

In contrast, the results presented in this work illustrate that competitive performance can be achieved through careful architectural design, domain-aware pre-training strategies, and efficient tokenization, without relying on excessive parameterization. This observation highlights a broader principle: in scientific modeling, model scale alone is not a reliable proxy for functional capability or practical utility. Evidence from benchmarks such as ProteinGym~\cite{pascalnotin_scaling_wall} suggests that, unlike language models trained on natural language, biological sequence models do not exhibit consistent scaling trends~\cite{scaling-law}. Model performance typically saturates and may even deteriorate when model size exceeds approximately 10B parameters. 

This divergence reflects fundamental differences between linguistic and biological data. Human language is highly structured and shaped by shared semantic conventions, whereas genomic sequences are inherently noisy, sparse, and shaped by evolutionary stochasticity. A substantial fraction of genetic variation is functionally neutral or redundant, creating background signal that can obscure meaningful biological constraints. In such settings, increasingly large models risk overfitting to statistical regularities that do not correspond to biological function~\cite{weinstein2022non}. These considerations suggest that progress in AI for the life sciences will depend not only on scaling existing architectures, but also on developing modeling strategies that are explicitly aligned with the structure and noise characteristics of biological data.

Within this context, GENERator exemplifies how biologically informed design choices can lead to models that are both effective and practical. By aligning data curation, tokenization, and architecture with genomic structure, GENERator achieves strong performance while remaining computationally accessible and interpretable. More broadly, this work highlights the importance of adapting machine learning methodologies to the specific properties of scientific domains, rather than directly transferring assumptions from natural language processing.

Finally, as part of this broader vision, the Gener Project is committed to advancing genomic research through open science and collaborative development. To promote transparency, reproducibility, and broad adoption, all data, code, and model weights associated with this work are released openly. We invite the research community to build upon these resources and to participate in collaborative efforts aimed at democratizing access to genomic foundation models and accelerating progress across functional genomics, metagenomics, and related fields.
\section{Methods}

\subsection{Data Preparation}
\label{sec:data_preparation}
For training GENERator, we sourced genomic DNA sequences from all eukaryotic organisms available in the RefSeq database. To investigate how training data composition influences representation learning and downstream generalization, we considered two alternative data preparation strategies that differ in whether pre-training is performed on gene-centric functional loci or on complete genomic sequence.

\paragraph{Functional sequence training.}  
In this setting, we construct a gene-centric pre-training corpus based on RefSeq annotations. Functional sequences are defined conservatively at the level of annotated \texttt{gene} entries in GenBank-format records~\cite{RefSeq}. For each annotated gene locus, we extract the genomic interval spanning the gene boundaries, which may include multiple coding regions (e.g., multiple CDS features across exons), intervening introns, untranslated regions (UTRs), and other transcribed regulatory segments associated with the gene. Where available in the annotation, promoter- and terminator-proximal regulatory regions are also included. This definition applies to both protein-coding genes and RNA genes, yielding gene-centric functional regions that are substantially longer than individual CDS, exon, or intron units. As a result, each training example naturally spans long genomic contexts, aligning with the long-context modeling objective of GENERator. The resulting functional corpus contains a total of 386 billion nucleotides.

\paragraph{All-sequence training.}  
As a baseline reflecting common practice in genomic language modeling, we also perform pre-training using all available eukaryotic genomic sequences from RefSeq without filtering by functional annotation. This all-sequence corpus comprises approximately 2 trillion nucleotides and includes both functional and non-functional genomic regions.

Empirically, all-sequence training achieves a lower pre-training loss (Figure~\ref{fig:pretrain_loss}), whereas functional sequence training consistently yields superior performance across downstream tasks (Section~\ref{sec:experiments}), including tasks in which non-functional genomic context is present at inference time. One plausible explanation for this counterintuitive behavior lies in the uneven distribution of semantic information across eukaryotic genomes. Unlike human language, where most tokens contribute meaningfully to semantic structure, genomic sequences contain large fractions of background noise that is weakly constrained or functionally redundant. Over evolutionary time, biologically functional sequence elements have emerged and been preserved by selection, resulting in localized regions with high information density, while much of the surrounding genomic sequence remains comparatively unconstrained~\cite{MLMSC, hemiplasy}. These evolutionarily constrained, gene-associated loci constitute the primary carriers of biological `semantics' in DNA.

In contrast, such functionally redundant regions are often dominated by low-complexity or repetitive patterns, including simple homopolymers (e.g., \texttt{AAAAAA}) and short tandem repeats (e.g., \texttt{GCGCGC})~\cite{dna-msa}. Such sequences are easy to model under next-token prediction objectives and can disproportionately reduce the training loss, despite contributing limited information relevant to biological function. Consequently, indiscriminate inclusion of large amounts of non-functional genomic background may bias the pre-training objective toward trivial statistical regularities, reducing the effectiveness of learned representations for downstream biological tasks.

This interpretation is further supported by population-genetic observations that mutation rates and tolerated variation are typically lower in gene-associated regions than in intergenic background~\cite{mutation-rates}, reflecting stronger purifying selection on functional sequence. Taken together, these considerations suggest that concentrating pre-training on gene-centric functional loci emphasizes evolutionarily meaningful patterns, leading to representations that generalize more effectively, even when evaluated on tasks involving broader genomic context.

\subsection{Tokenization}
\label{sec:tokenization}
This section describes the tokenization strategies considered for converting genomic DNA sequences into inputs suitable for language modeling. There are three classes of tokenizers commonly used in genomic sequence modeling:
\begin{itemize}
    \item \textbf{Single-nucleotide tokenization.} Each nucleotide (\texttt{A}, \texttt{T}, \texttt{C}, \texttt{G}) is treated as an individual token, providing the highest possible sequence resolution. This scheme is adopted by models such as Evo, Caduceus, HyenaDNA, and megaDNA.
    \item \textbf{$k$-mer tokenization.} Consecutive substrings of length $k$ are treated as atomic tokens. This approach is used by models such as DNABERT and the Nucleotide Transformer.
    \item \textbf{Byte pair encoding (BPE).} Originally developed for natural language processing, BPE learns a vocabulary of frequently occurring substrings by iteratively merging character pairs~\cite{SentencePiece}. Genomic models such as DNABERT-2, GROVER and GenomeOcean adopt this strategy.
\end{itemize}

The choice of tokenization reflects a fundamental trade-off between sequence resolution and contextual coverage. Single-nucleotide tokenization preserves fine-grained local detail, but under a fixed token budget results in a substantially shorter effective context length than $k$-mer or BPE tokenization. In transformer architectures, where the computational cost of self-attention scales quadratically with token sequence length, this limitation becomes particularly restrictive. Unlike protein sequences, which typically span hundreds to a few thousand amino acids and can be modeled directly using standard transformer architectures~\cite{esm2}, genomic regions of interest often extend to tens of thousands of nucleotides, placing severe constraints on feasible input lengths.

A variety of architectural strategies have been proposed to mitigate this tension under single-nucleotide tokenization. Several models augment sequence modeling with Conv1D or convolution-like operators to increase receptive fields, including Enformer~\cite{enformer}, HyenaDNA~\cite{HyenaDNA}, Evo~\cite{Evo}, Evo2~\cite{Evo2}, and AlphaGenome~\cite{alphagenome}. In parallel, state space models (SSMs), such as Caduceus~\cite{Caduceus} and HybriDNA~\cite{hybridna}, replace explicit attention with recurrent-style hidden state updates that scale linearly with sequence length. Although these approaches can process much longer nominal input sequences, our experimental results (Section~\ref{sec:kmer_predition}) show that increased nominal context length alone does not necessarily translate into improved long-range sequence understanding.

Based on these observations, we focus on extending effective context coverage within the transformer framework through principled tokenization choices. Previous studies such as DNABERT-2~\cite{DNABERT-2} and GROVER~\cite{GROVER} examined tokenization strategies in the context of masked language modeling and reported advantages for BPE. In contrast, the impact of tokenization on autoregressive (next-token prediction) genomic language models has not been systematically explored. Although BPE is highly effective in natural language processing, our empirical results indicate that $k$-mer tokenization substantially outperforms both single-nucleotide and BPE tokenization for causal genomic language modeling. A detailed empirical comparison is provided in Section~\ref{sec:kmer_predition}, where the 6-mer tokenizer emerges as the most effective configuration.

\subsection{Pre-training}
This section describes the pre-training procedure for GENERator, which uses the functional sequence training strategy (Section~\ref{sec:data_preparation}) together with 6-mer tokenization. GENERator adopts a decoder-only transformer architecture based on the LLaMA framework~\cite{llama}, with full architectural specifications provided in Table~\ref{tab:model_arch}. Pre-training is performed with a global batch size of 2 million tokens using the AdamW optimizer~\cite{adamw} and a cosine learning rate schedule with a warm-up phase.

The model is trained for six epochs on a corpus of 386 billion nucleotides; with 6-mer tokenization, this corresponds to a total of 386 billion tokens processed during training. To account for phase offsets introduced by $k$-mer tokenization, we randomize the starting offset of each training sequence uniformly between 0 and 5 nucleotides. As a result, the model observes different 6-mer segmentations of the same underlying nucleotide corpus across training, reducing sensitivity to tokenization boundaries and improving robustness to sequence alignment variation. 

Efficient long-context pre-training is enabled through the use of FlashAttention~\cite{flashattention2} and the ZeRO optimizer~\cite{deepspeed} to improve memory and communication efficiency. Additional implementation details are provided in Section~\ref{sec:pretrain_supp}.

\subsection{Downstream Tasks}
The downstream tasks evaluated in this study are used to assess both the generalization capability and practical utility of genomic language models. These tasks span a range of objectives, beginning with zero-shot probing tasks that evaluate intrinsic representational and generative capacities without any parameter updates, followed by fine-tuning tasks that assess adaptability to specific biological objectives. This section describes the methodological framework for each task, focusing on problem formulation, data construction, and model usage. Experimental results and comparative analyses are presented in Section~\ref{sec:experiments}.

\subsubsection{Embedding Representation}
\label{sec:embedding_clustering}
To evaluate the intrinsic quality and biological relevance of the sequence representations learned by GENERator, we perform unsupervised embedding clustering analysis. This analysis tests whether the model’s latent sequence embeddings capture fundamental taxonomic structure directly from raw genomic DNA, without fine-tuning or task-specific supervision. 

We constructed an evaluation dataset by sampling genomic DNA fragments from six major eukaryotic taxonomic groups defined in the RefSeq database: \emph{protozoa, fungi, plant, invertebrate, mammalian, and vertebrate (other)}. For each group, species were randomly selected, and from each species a fixed-length genomic segment was extracted at a random genomic location. Each sequence was then processed by the pre-trained model, and the final hidden state corresponding to the last token was used as a fixed-dimensional embedding representing the entire input sequence.

To visualize the resulting high-dimensional embeddings and assess their structure, we applied Uniform Manifold Approximation and Projection (UMAP)~\cite{umap} to reduce the embeddings to two dimensions. Clustering quality was evaluated by examining the separation and cohesion of embeddings corresponding to the six taxonomic groups in the low-dimensional space.

This unsupervised analysis provides a baseline assessment of whether GENERator learns generalizable, biologically meaningful representations from genomic sequence alone, which is a prerequisite for downstream supervised and generative applications.

\subsubsection{Sequence Recovery}
\label{sec:sequence_recovery}
To compare the intrinsic generative capabilities of genomic language models, a natural approach is to evaluate their pre-training loss on a standardized test set. However, such comparisons are fundamentally confounded by differences in tokenization. For instance, a model using a 6-mer tokenizer faces a 4096-way classification problem, whereas one using single-nucleotide tokenization faces only a 4-way classification. These inherent disparities in prediction space complexity render raw loss values incomparable across models with different tokenizers (Figure~\ref{fig:pretrain_loss_tokenizer}).

To address this limitation, we introduce \emph{sequence recovery} as a unified, tokenization-agnostic evaluation framework for generative DNA models. A related evaluation concept, termed \emph{next $k$-mer prediction}, was previously introduced in GROVER~\cite{GROVER}, where it was applied after task-specific fine-tuning. In contrast, we formulate sequence recovery as a zero-shot evaluation that can be applied uniformly to a wide range of pre-trained models without additional parameter updates.

In sequence recovery, a model is prompted with a genomic sequence segment and tasked with generating the subsequent nucleotides in an autoregressive manner. For masked language models, which are not natively autoregressive, we implement a sequential decoding protocol in which a \texttt{<mask>} token is appended to the input at each generation step, and the predicted token is incorporated into the input for the next step. Generated sequences are then compared to the reference sequence, and recovery accuracy is quantified by nucleotide-level overlap.

The evaluation dataset is constructed to reflect realistic generative scenarios in genomics. Generated regions are restricted to gene-centric functional regions, consistent with the goal of producing biologically meaningful DNA sequences. At the same time, input segments are selected to include substantial non-gene context, ensuring that models trained on all-sequence data can leverage broader genomic information when available.

Beyond benchmarking, sequence recovery also has practical relevance for genomic analysis. For example, it can be applied to gap filling in genome assembly, where repetitive or complex regions are often missing, as well as to post hoc correction of sequencing errors such as substitutions or small indels. These use cases illustrate that sequence recovery provides not only a principled evaluation metric, but also a practical tool for downstream genomic applications.

\subsection{Variant Effect Prediction}
\label{sec:variant-effect-prediction}
As a zero-shot application of GENERator, we introduce an alignment-free approach for variant effect prediction (VEP), which aims to estimate the potential functional impact of single nucleotide variants (SNVs) based solely on the learned representation of DNA sequences. 
The underlying assumption is that large-scale pre-training induces a probabilistic prior over evolutionarily conserved sequence patterns. Under this assumption, nucleotides consistent with the learned prior are assigned higher probability, whereas deviations from this prior—such as potentially deleterious mutations—are expected to receive lower probability.

Formally, given a reference sequence $S = (s_1, s_2, \dots, s_i, \dots, s_L)$ and a single nucleotide variant at position $i$, where the reference allele is $R$ and the alternative allele is $A$, we define the VEP score as the log-likelihood ratio
\begin{equation*}
    \text{VEP}(R \rightarrow A) = \log \frac{p(s_i = R \mid S_{\setminus i})}{p(s_i = A \mid S_{\setminus i})},
\end{equation*}
where $S_{\setminus i}$ denotes the full sequence context excluding position $i$. Positive VEP scores indicate that the reference allele is preferred over the alternative allele under the model, consistent with stronger evolutionary constraint at that position.

A practical challenge arises from the use of $k$-mer tokenization. Directly comparing the probabilities of reference and alternative $k$-mer tokens involves a 4096-way classification space for 6-mer models, whereas the biological interpretation of interest is defined at single-nucleotide resolution. Token-level probabilities are therefore both difficult to compare across alleles and sensitive to numerical scale. To address this mismatch, we project token-level probabilities onto nucleotide-level probabilities through marginalization.

Specifically, for a $k$-mer token $t$ that spans position $i$, we compute the probability that nucleotide $X \in \{\texttt{A}, \texttt{T}, \texttt{C}, \texttt{G}\}$ occurs at position $i$ as
\begin{equation*}
    p(s_i = X) = \sum_{t \in \mathcal{T}_{i}} p(t)\,\mathbb{I}(t_j = X),
\end{equation*}
where $\mathcal{T}_{i}$ denotes the set of $k$-mer tokens covering position $i$, $j$ is the relative offset of position $i$ within token $t$, and $\mathbb{I}$ is the indicator function. This marginalization reduces the problem from a 4096-way to a 4-way probability comparison, yielding scores that are both numerically stable and directly interpretable at single-nucleotide resolution.

Conceptually, this approach is related to classical conservation-based methods such as phyloP~\cite{phylop} and CADD~\cite{cadd}, which assess evolutionary tolerance at individual genomic positions. However, those methods rely on multiple sequence alignments and curated comparative genomic resources, which are computationally expensive and often unavailable outside of well-studied organisms. In contrast, the approach described here is entirely alignment-free and requires only a single reference sequence, illustrating how generative genomic language models can capture biologically meaningful constraints through zero-shot inference.

\subsubsection{Benchmark Evaluations}
\label{sec:benchmark}
In addition to zero-shot evaluations of intrinsic model capabilities, we assess performance on a range of supervised benchmarks that require task-specific fine-tuning. These benchmarks are used to evaluate the model’s ability to adapt to concrete biological applications.

We consider two widely used benchmark suites: Genomic Benchmarks~\cite{genomic-benchmarks} and NT tasks~\cite{nucleotide-transformer}. Genomic Benchmarks are centered on human genomic data, with additional datasets from organisms such as mouse, roundworm, and fruit fly. The benchmark includes datasets for regulatory element classification (e.g., promoters and enhancers), as well as \emph{demo} datasets for species or transcript-type classification and \emph{dummy} datasets intended for rapid prototyping. In contrast, NT tasks span broader species diversity and include promoter and enhancer classification, yeast epigenetic mark prediction, and splice site identification across more than 100 organisms. While these benchmarks are well established, the sequences they contain are predominantly short, typically on the order of hundreds of nucleotides, which limits their coverage of long-range genomic context.

To complement these existing benchmarks, we introduce two additional sequence classification tasks designed to probe model behavior across longer genomic contexts and broader taxonomic structure. The \emph{gene classification} task evaluates the model’s ability to discriminate among gene-associated sequence types using sequences ranging from 100 to 5,000 base pairs. This task includes six gene categories (\emph{CDS, pseudo, tRNA, rRNA, ncRNA, miscRNA}), together with \emph{control} sequences sampled from intergenic regions. Samples are drawn in a balanced manner from six major eukaryotic taxonomic groups in RefSeq: \emph{protozoa, fungi, plant, invertebrate, mammalian, and vertebrate (other)}. The classification objective is to predict the gene type.

The \emph{taxonomic classification} task is designed to assess model performance on substantially longer sequences that contain a mixture of gene and non-gene regions. In this task, sequences are standardized to 96k base pairs and are similarly balanced across the same six eukaryotic taxonomic groups. The objective is to predict the taxonomic group from which each sequence is derived.

\subsubsection{Central Dogma}
\label{sec:central_dogma}
The central dogma of molecular biology describes the flow of genetic information from DNA to RNA and ultimately to proteins, which constitute the primary functional components of living systems~\cite{central-dogma}. As the fundamental carrier of biological information, DNA encodes the instructions for RNA synthesis and protein production. A genomic language model that captures biologically meaningful structure should therefore reflect not only nucleotide-level sequence patterns, but also constraints associated with downstream protein-coding and folding processes.

To assess this capability, we design a task grounded in the central dogma by fine-tuning GENERator on protein-coding DNA sequences and evaluating its ability to generate novel DNA sequences that can be translated into structurally plausible proteins. Protein-coding DNA sequences from two well-characterized and evolutionarily distinct protein families, cytochrome P450 and histone, were curated by cross-referencing protein entries in UniProt~\cite{UniProt} with corresponding gene annotations in the RefSeq~\cite{RefSeq} database, ensuring consistency between nucleotide sequences and their translated protein products. GENERator was then fine-tuned on these family-specific datasets to generate sequences with comparable coding characteristics. Generated DNA sequences were evaluated using the following criteria; results are reported in Section~\ref{sec:central_dogma_exp}.

\paragraph{Translation validity and coding integrity.}
Generated DNA sequences were translated into amino acid sequences using the standard genetic code. Translation success was evaluated by examining whether generated sequences formed continuous and stable open reading frames, without premature stop codons or frameshift disruptions. Length distributions of translated protein sequences were compared with those of natural protein sequences from the same family, providing a basic check of coding completeness and structural consistency at the DNA level.

\paragraph{Protein sequence plausibility.}
To evaluate sequence-level plausibility in protein space, translated amino acid sequences were scored using the protein language model ProGen2~\cite{progen2}. Perplexity values of generated proteins were compared against those of natural family members. In addition, a control set was constructed by randomly permuting amino acid identities while preserving the length distribution of natural sequences. This comparison assesses whether generated proteins are recognized by a protein language model as protein-like sequences rather than random amino acid strings.

\paragraph{Structure prediction confidence.}
Three-dimensional protein structures were predicted for generated sequences using AlphaFold3~\cite{AlphaFold3}. Prediction confidence was assessed using pLDDT scores, which quantify the model’s confidence in the predicted structures. This step evaluates whether generated sequences can be folded into stable protein structures with high confidence by a state-of-the-art structure prediction model.

\paragraph{Structural similarity and novelty assessment.}
Predicted protein structures were compared against experimentally resolved structures in the Protein Data Bank (PDB)~\cite{RCSBPDB} using FoldSeek~\cite{Foldseek}. Structural similarity was quantified using TM-score to identify natural proteins with similar folds. For matched structures, sequence identity between generated proteins and their closest structural neighbors was also computed. This joint analysis of structure-level similarity and sequence-level divergence allows assessment of whether generated proteins recapitulate known protein folds without trivial sequence copying.

\subsubsection{Cis-Regulatory Element Design}
\label{sec:sequence_design}
This section describes the methodological framework used for cis-regulatory element (CRE) design with GENERator. CREs broadly encompass non-coding regulatory sequences such as enhancers, promoters, and silencers. 

In this study, we focus on CRE design with an emphasis on enhancer activity, while allowing the model to explore both activating and repressive regulatory regimes. This focus is motivated by the central role of enhancers in transcriptional regulation and by the availability of large-scale, quantitative activity measurements in the DeepSTARR dataset~\cite{DeepSTARR}, which spans a broad spectrum of regulatory activity.

Our CRE design framework consists of two complementary components derived from the same underlying model: an activity-conditioned CRE generator that enables prompt-guided sequence design, and a CRE activity predictor that provides fine-grained quantitative scoring for candidate ranking and interpretation. The generator provides coarse-grained control over regulatory activity at generation time, while the predictor refines this control by enabling continuous scoring and selection of candidate sequences.

\paragraph{Activity-conditioned CRE generation.}
To enable conditional generation, we discretize DeepSTARR activity labels into three regimes based on activity-score quantiles: the top quartile (high activity), the middle 50\% (medium activity), and the bottom quartile (low activity). Each regime is associated with a dedicated prefix token (\texttt{<high>}, \texttt{<mid>}, \texttt{<low>}). GENERator is fine-tuned on the resulting token-labeled sequences so that, during inference, prompting the model with a prefix token conditions generation toward the corresponding activity regime. This activity-conditioned generator provides coarse-grained, prompt-guided control over CRE activity.

\paragraph{CRE activity prediction.}
In parallel, we fine-tune GENERator as a supervised CRE activity predictor that maps input sequences to continuous activity scores. This predictor is trained on the same DeepSTARR dataset and evaluated on the held-out test set to assess sequence-to-activity modeling performance (Table~\ref{tab:enhancer_benchmark}). The predictor serves two roles: (i) computational evaluation of model understanding of regulatory sequence grammar, and (ii) fine-grained ranking of generated sequences within each activity regime.

\paragraph{Predictor-guided candidate selection and experimental validation.}
CRE sequences generated under each prefix condition are scored by the CRE activity predictor to enable within-group ranking. For experimental validation, candidate sequences are selected from the top of the \texttt{<high>} group, the bottom of the \texttt{<low>} group, and randomly sampled from the \texttt{<mid>} group to provide a matched control. Selected sequences are synthesized and evaluated using UMI-STARR-seq assays in \textit{D.\ melanogaster} S2 cells; experimental procedures and results are described in Section~\ref{sec:enhancer_design_exp}.

\paragraph{Base-level contribution scores.}
Beyond sequence-level activity prediction, we further analyze the CRE activity predictor at single-nucleotide resolution to identify positions that contribute most strongly to predicted regulatory activity. To this end, we compute base-level contribution scores using an in silico mutagenesis strategy. Given a CRE sequence $S = (s_1, s_2, \dots, s_i, \dots, s_L)$ and a trained predictor $f(\cdot)$, the contribution of position $i$ is evaluated by substituting the nucleotide at that position with each of the four possible bases while keeping all other positions fixed. Let $S_{i \rightarrow x}$ denote the sequence obtained by replacing the nucleotide at position $i$ with base $x \in \{\texttt{A}, \texttt{T}, \texttt{C}, \texttt{G}\}$. The contribution score at position $i$ is defined as
\begin{equation*}
    \mathcal{C}(i)
    \;=\;
    f\!\left(S\right)
    \;-\;
    \frac{1}{3}
    \sum_{\substack{x \in \{\texttt{A}, \texttt{T}, \texttt{C}, \texttt{G}\} \\ x \neq s_i}}
    f\!\left(S_{i \rightarrow x}\right),
\end{equation*}
where $s_i$ denotes the original nucleotide at position $i$. This score measures the difference between the predicted activity of the original sequence and the mean predicted activity under the three alternative single-nucleotide substitutions, which provides a position-specific local baseline. The resulting position-wise contribution profiles are used for downstream analyses, including motif overlap and identification of high-contribution regulatory regions.

Together, this framework couples activity-conditioned CRE generation, predictor-guided candidate ranking, high-throughput experimental validation, and single-nucleotide interpretation. The corresponding computational benchmarks and experimental results are reported in Section~\ref{sec:enhancer_design_exp}.

\newpage

\section*{Acknowledgements}
The authors thank Guojie Zhang, Guangji Chen, Guangyong Chen, and Qinqin Long for their early support and helpful discussions during the initial stages of the project. We also thank Chuan Cao, Haiguang Liu, and Tao Qin for their support and insightful discussions during the final phase of the work. In addition, we thank Lulu Zhang for her assistance with administrative coordination and management throughout the project.

\section*{Author Contributions}
Q.L. was the primary lead of the project and contributed to study conception, data preparation, model design, downstream analyses, and manuscript drafting and revision. W.W. made major technical contributions to model development, downstream evaluation, and open-source code implementation. Y.Z. performed CRE synthesis and validation experiments, conducted data analysis and visualization, and contributed to drafting the corresponding sections. Z.Z. carried out computational evaluations, particularly related to Evo2 benchmarking, and contributed to discussions.

Y.L. conceived and supervised the CRE design framework and oversaw the wet-lab experiments. R.C., J.Q., Y.B., and C.W. contributed to CRE-related data processing and analysis. M.L., K.F., Y.Zhu, and Z.Zhang contributed to discussions and manuscript drafting.

J.Y., F.F., and J.T. provided supervision, computational resources, and insightful discussions. H.X. and Z.W. provided overall project supervision, strategic guidance, and computational and personnel support. All authors reviewed and approved the final manuscript.

\section*{Competing Interests}
The authors declare no competing interests.

\section*{Data Availability}

The raw UMI-STARR-seq sequence data generated in this study, together with the corresponding experimental activity measurements, have been deposited in the Genome Sequence Archive (GSA) at the BIG Data Center (\url{https://bigd.big.ac.cn/}) under the accession code \texttt{PRJCA056161}.

The GENERator model weights, as well as the associated training and benchmarking datasets, are publicly available at \url{https://huggingface.co/GenerTeam}.

\section*{Code Availability}

All Python scripts used for downstream analyses are available on GitHub at \url{https://github.com/GenerTeam}.

\bibliographystyle{plainnat}
\bibliography{references}

\setcounter{figure}{0}
\setcounter{table}{0}
\setcounter{equation}{0}
\renewcommand{\thefigure}{S\arabic{figure}}
\renewcommand{\thetable}{S\arabic{table}}

\newpage
\appendix
\section{Supplementary Experiment Results}

\subsection{Zero-shot Evaluations}

\begin{figure}[ht]
    \centering
    \includegraphics[width=\textwidth]{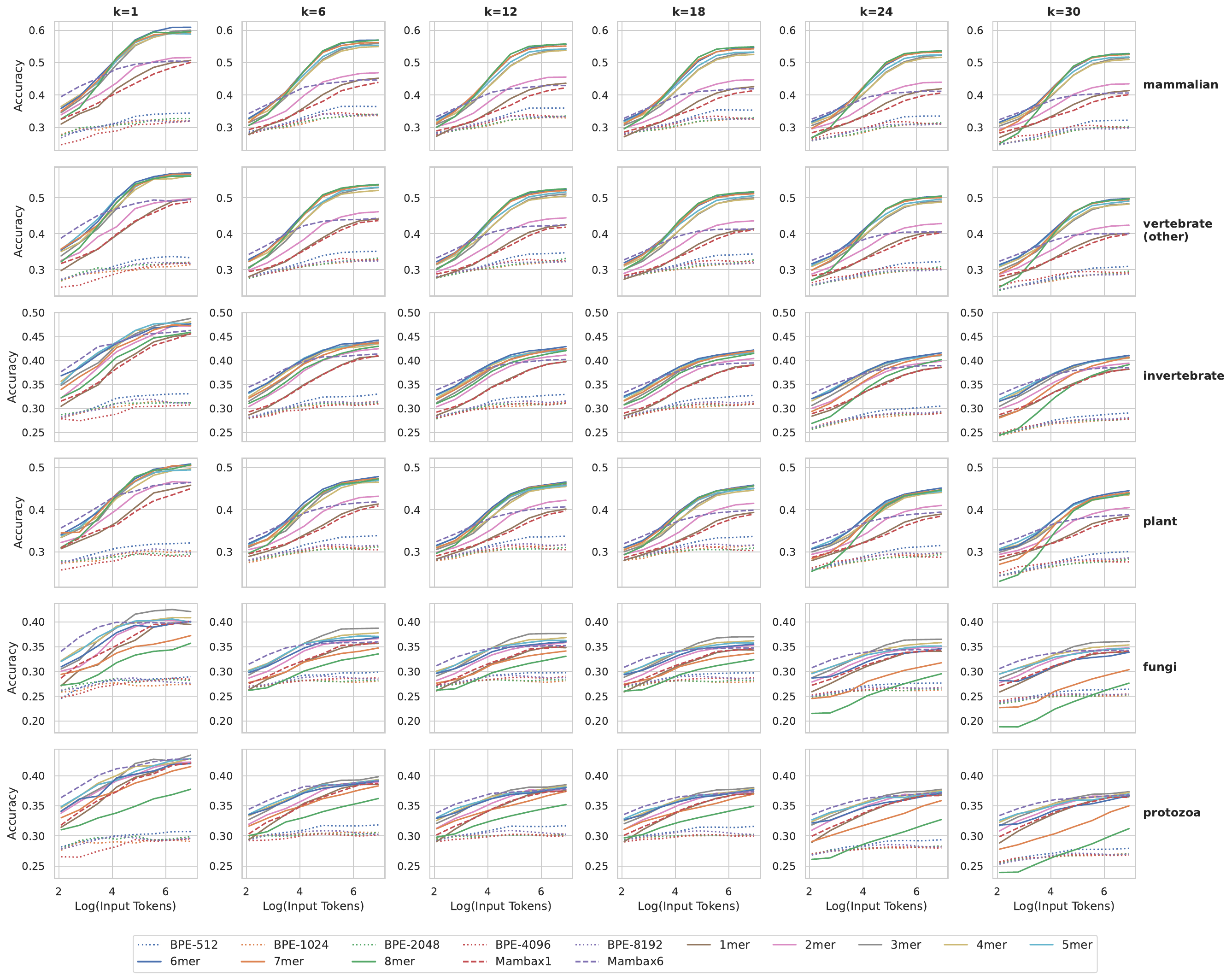}
    \caption{Comparative sequence recovery accuracy across tokenization strategies and model architectures. Each cell shows accuracy (y-axis) versus input tokens (x-axis, 8-1024) for a taxonomic group and prediction length (1-30 bp). Lines represent: BPE tokenizers (all vocabulary sizes, consistently poor), $k$-mer tokenizers with $k$=1-2 (moderate performance), $k$-mer tokenizers with $k$=3-8 (superior performance), and Mamba architectures (Mamba×1: same tokens as 1-mer; Mamba×6: same bp as 6-mer). While $k$=3-8 perform similarly well, 6-mer achieves optimal balance - smaller $k$ values ($k$=3-5) provide limited context extension, while larger $k$ (7,8) suffer in data-scarce groups (fungi, protozoa) and with fewer tokens. Mamba architectures reveal critical insights: both Mamba×1 and Mamba×6 achieve only 1mer-level accuracy despite the 6-fold longer context in Mamba×6. Mamba×6 shows initial advantage with few tokens (effectively more tokens) but is surpassed as token count increases. The convergence of Mamba×1 (1024 tokens) and Mamba×6 (6144 tokens) at similar accuracy indicates that while SSMs can process long inputs, they lack proportional gains in contextual understanding compared to transformers.}
    \label{fig:kmer_tokenizer}
\end{figure}

\begin{figure}[ht]
    \centering
    \includegraphics[width=\textwidth]{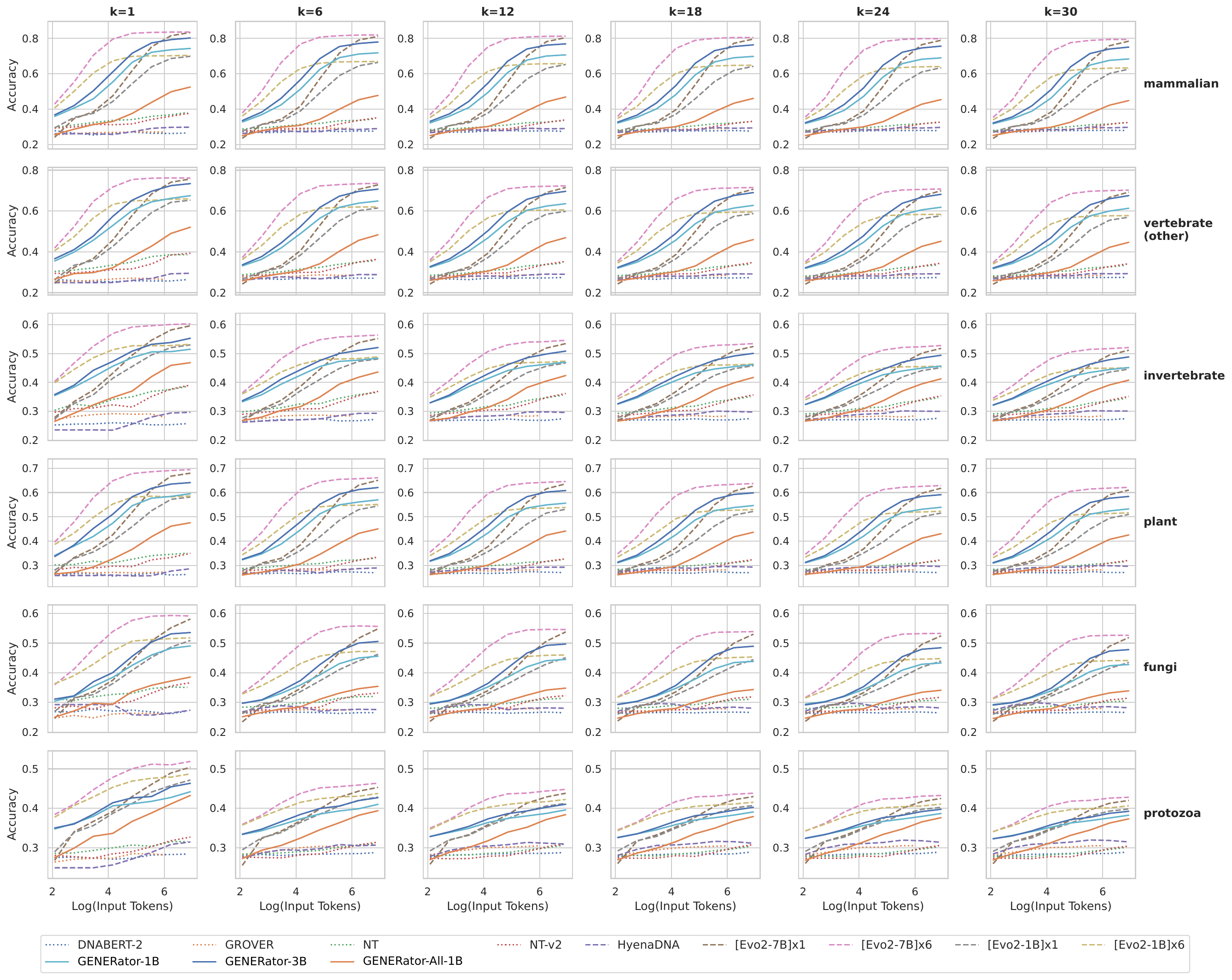}
    \caption{Comparative sequence recovery accuracy across genomic foundation models. Each cell displays accuracy (y-axis) versus input tokens (x-axis, 8-1024) for a taxonomic group and prediction length (1-30 bp). Lines represent various baseline models: MLM models (consistently poor as expected due to lack of generative capability); HyenaDNA (autoregressive but severely limited by its 55M parameters, a capacity highly disproportionate to its 1M context length); and the leading autoregressive models GENERator and Evo2. We compare two Evo2 configurations: [Evo2]×1 (same token count as other models, representing comparable efficiency) and [Evo2]×6 (same nucleotide count as GENERator, but with inference times several to dozens of times slower than GENERator). Evo2 series reveal patterns similar to Mamba-2 in Figure \ref{fig:kmer_tokenizer}: [Evo2]×6 shows initial advantage with few tokens (more effective tokens) but this advantage diminishes rapidly within a few hundred tokens. Critically, [Evo2]×1 (1024 tokens) and [Evo2]×6 (6144 tokens) achieve nearly identical accuracy, indicating ineffective utilization of extended context despite the long-context capabilities claimed by Evo2. While Evo2-7B achieves higher absolute accuracy (attributable to larger training data, more steps, and greater parameters), GENERator demonstrates superior parameter efficiency: GENERator-1B surpasses Evo2-1B, and GENERator-3B approaches Evo2-7B performance. Notably, GENERator maintains a clear upward accuracy trend even at the maximum tested length (1024 tokens, 6144 bp), suggesting better long-context utilization.}
    \label{fig:kmer_model}
\end{figure}

\paragraph{Sequence Recovery}
Comprehensive evaluation of sequence recovery performance across taxonomic groups is provided in Figure~\ref{fig:kmer_tokenizer} and Figure~\ref{fig:kmer_model}. Our analysis examines accuracy across varying prediction lengths and input token lengths to assess contextual understanding. Figure~\ref{fig:kmer_tokenizer} compares tokenization strategies, confirming the consistent advantage of 6-mer tokenization across all experimental conditions. Figure~\ref{fig:kmer_model} demonstrates that both GENERator series and Evo series significantly outperform all other baseline models, with GENERator-1B surpassing Evo2-1B at comparable parameter scales and GENERator-3B achieving competitive performance against Evo2-7B. Notably, GENERator achieves orders-of-magnitude higher throughput than Evo2 models, demonstrating substantially improved computational efficiency. For generation time benchmarking, both GENERator and Evo2 series (excluding Evo2-40B) were evaluated under uniform conditions on L40S GPUs, measuring the time to generate 512 bp from a 6144 bp input sequence, with dtype settings preserved from original implementations (FP8 for Evo2 series, BF16 for GENERator series).

\paragraph{Variant Effect Prediction}
Table~\ref{tab:variant_effect_prediction} presents zero-shot variant effect prediction performance on the ClinVar dataset. The GENERator series achieves competitive performance among sequence-only models, with GENERator-3B matching the best-performing Evo2-7B model in AUROC (0.921).

\subsection{Task-specific Fine-tuning}
\paragraph{Benchmark Tasks}
Complete benchmark evaluation results are provided in the following tables: Revised NT tasks (Table~\ref{tab:nucleotide_transformer_tasks_revised}), Original NT tasks (Table~\ref{tab:nucleotide_transformer_tasks}), Genomic Benchmarks (Table~\ref{tab:genomic_benchmarks}), and Gener tasks (Table~\ref{tab:gener_tasks}). Reported values represent accuracy averaged over 10-fold cross-validation, with standard errors indicated in parentheses. Benchmark scores are sourced directly from the Nucleotide Transformer paper~\cite{nucleotide-transformer} where applicable; remaining models were evaluated uniformly following the experimental configuration detailed in Section~\ref{sec:exp_config_supp}. Specific model implementations and hyperparameter configurations are documented in Tables~\ref{tab:benchmark_hyperparam1} to \ref{tab:benchmark_hyperparam4}.

\begin{figure}[ht]
    \centering
    \includegraphics[width=0.9\textwidth]{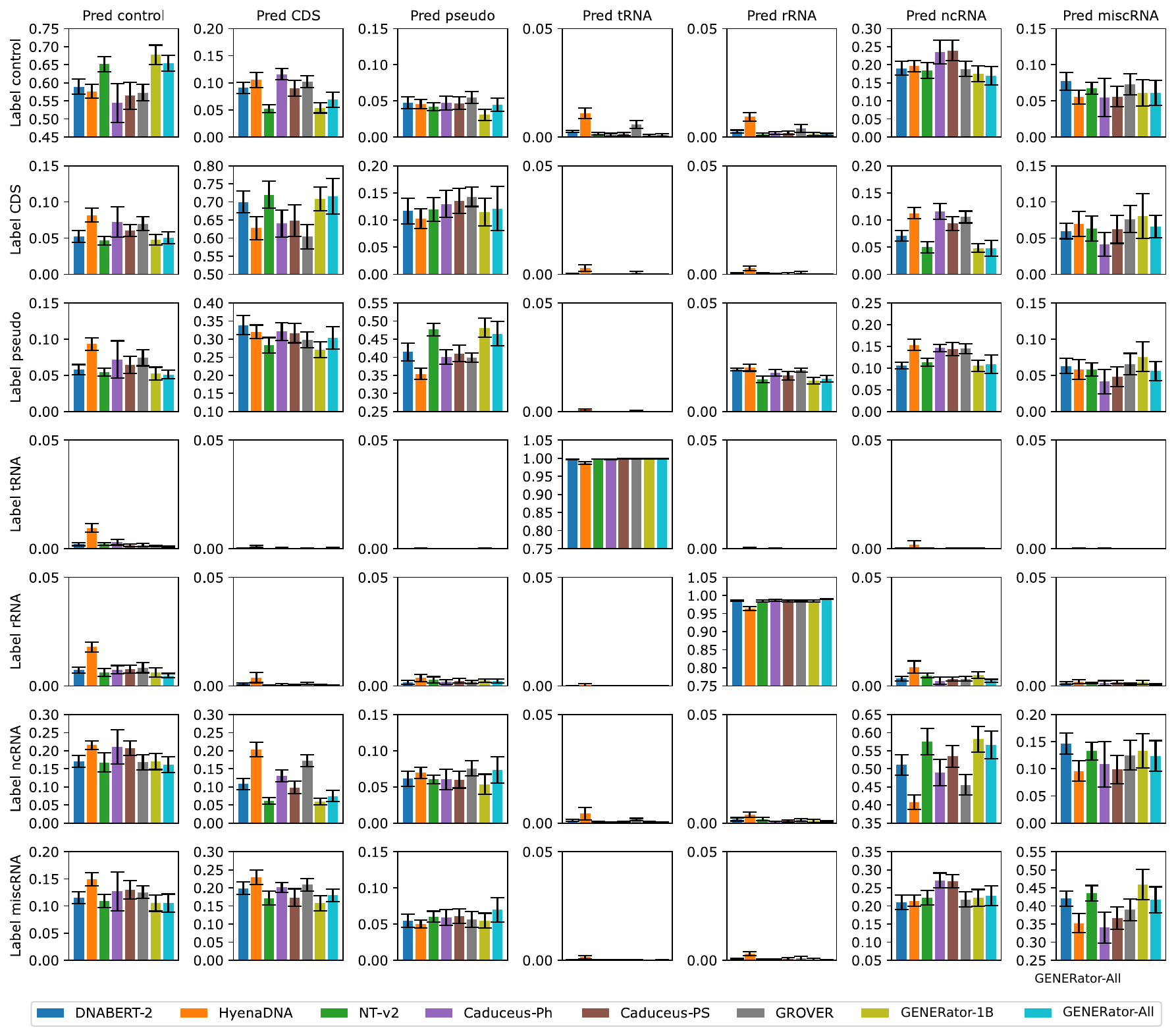}
    \caption{Confusion matrices for the gene classification task. The task involves predicting gene types from DNA sequences across six categories (CDS, pseudo, tRNA, rRNA, ncRNA, miscRNA) and a control group (intergenic). Each cell represents a bar plot comparing model performance for a specific label-prediction pair. In diagonal cells (correct predictions), taller bars indicate better performance; in off-diagonal cells (misclassifications), shorter bars are desirable. GENERator demonstrates top-tier accuracy across nearly all gene types.}
    \label{fig:gene_classification}
\end{figure}

\begin{figure}[ht]
    \centering
    \includegraphics[width=0.9\textwidth]{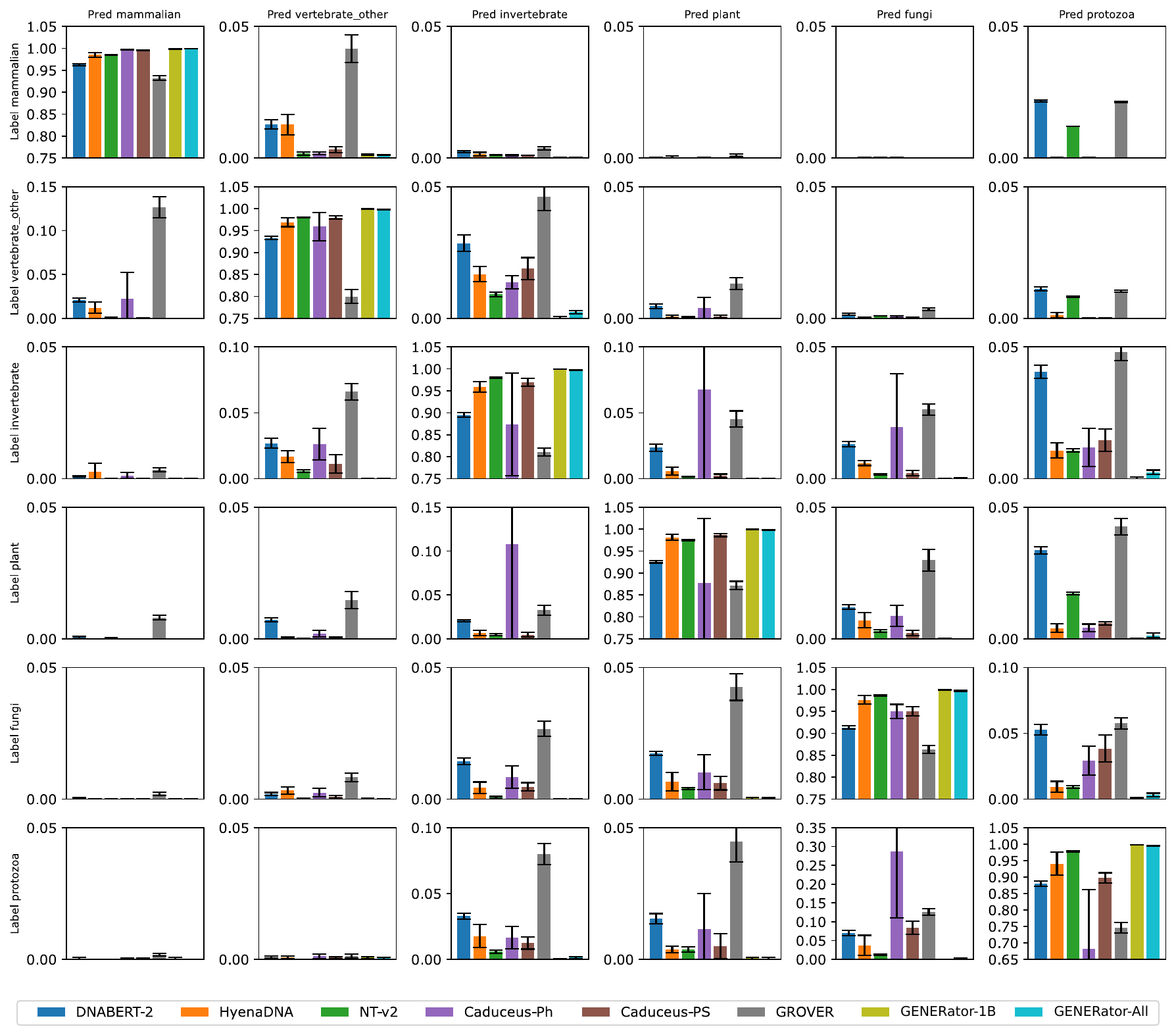}
    \caption{Confusion matrices for the taxonomic classification task. The task requires classifying 96k bp genomic sequences into six eukaryotic taxonomic groups: protozoa, fungi, plant, invertebrate, mammalian, and vertebrate (other). Each cell displays a bar plot of model predictions, with bar colors representing different models. Diagonal cells (correct taxonomic assignments) show GENERator achieving consistently high accuracy across all groups, while off-diagonal cells reveal minimal confusion between distant taxonomic groups.}
    \label{fig:species_classification}
\end{figure}

\paragraph{Gener Tasks Evaluation}
Detailed results for the gene classification and taxonomic classification tasks are presented in Figure~\ref{fig:gene_classification} and Figure~\ref{fig:species_classification}, respectively. The GENERator achieves superior average performance and maintains consistent accuracy across diverse gene categories and taxonomic groups, demonstrating robust generalization capability. These tasks evaluate the model's ability to capture functional and evolutionary information encoded in genomic sequences.

\paragraph{CRE Activity Prediction}
Table~\ref{tab:enhancer_benchmark} summarizes enhancer CRE prediction performance on the DeepSTARR hold-out test set following model fine-tuning. GENERator achieves the highest Pearson correlation coefficients for both developmental (Dev, 0.71) and housekeeping (Hk, 0.80) promoters, outperforming both the original DeepSTARR model and Nucleotide Transformer baselines.

\paragraph{CRE Design and Validation}
Quality control and experimental validation metrics for the UMI-STARR-seq assays are summarized in Supplementary Figures~\ref{fig:enhancer_supp_1}--\ref{fig:enhancer_supp_3}. In the input libraries, 99.64\% and 99.84\% of synthesized oligonucleotides were detected for the developmental (Dev) and housekeeping (Hk) conditions, respectively, with median UMI counts of 214 (Dev) and 341 (Hk), indicating high library complexity (Figure~\ref{fig:enhancer_supp_1}A). Biological replicates exhibited strong reproducibility, with $R^2 \geq 0.94$ (Figure~\ref{fig:enhancer_supp_1}B). Measured CRE activities spanned wide dynamic ranges, with $\log_2$ fold-change values from $-10.57$ to $9.97$ for Dev CREs and from $-11.73$ to $9.41$ for Hk CREs (Figure~\ref{fig:enhancer_supp_2}A). Supplementary Figure~\ref{fig:enhancer_supp_2} further reports activity distributions and prediction correlations for GENERator-designed CREs, while Supplementary Figure~\ref{fig:enhancer_supp_3} presents comparative analyses with prior design approaches and transcription factor motif enrichment across activity categories.

\section{Pre-training Details}
\label{sec:pretrain_supp}
\paragraph{Model Architecture and Hyperparameters}
The complete model architecture specifications are provided in Table~\ref{tab:model_arch}. We employed the AdamW optimizer~\cite{adamw} with $\beta_1 = 0.9$, $\beta_2 = 0.95$, and weight decay of $0.1$. The learning rate schedule combined linear warm-up and cosine decay: rates increased linearly from zero to a peak value of $4 \times 10^{-4}$ over the first 2,000 steps, then followed cosine decay to 10\% of the peak value by training completion. Gradient clipping was applied with a maximum norm of $1.0$.

\paragraph{Training Configuration}
Following standard large language model pre-training practices, we used a batch size of 2 million tokens. With a maximum sequence length of 16,384 tokens, this configuration yielded 128 samples per batch. Each sample was initialized with random starting positions between 0 and 5 at every epoch to enhance robustness. The complete pre-training spanned 6 epochs, corresponding to approximately 185,000 training steps.

\paragraph{Computational Efficiency}
Training efficiency was optimized using Flash Attention~\cite{flashattention,flashattention2} and Zero Redundancy Optimizer~\cite{deepspeed,fsdp}. The pre-training process for the 1B parameter model required 11,776 A100-hours (368 hours on 32 GPUs) with an MFU of 46\%. The 3B model maintained this efficiency at scale, requiring approximately 2.5$\times$ the computational budget.

\begin{figure}[ht]
    \centering
    \includegraphics[width=0.7\textwidth]{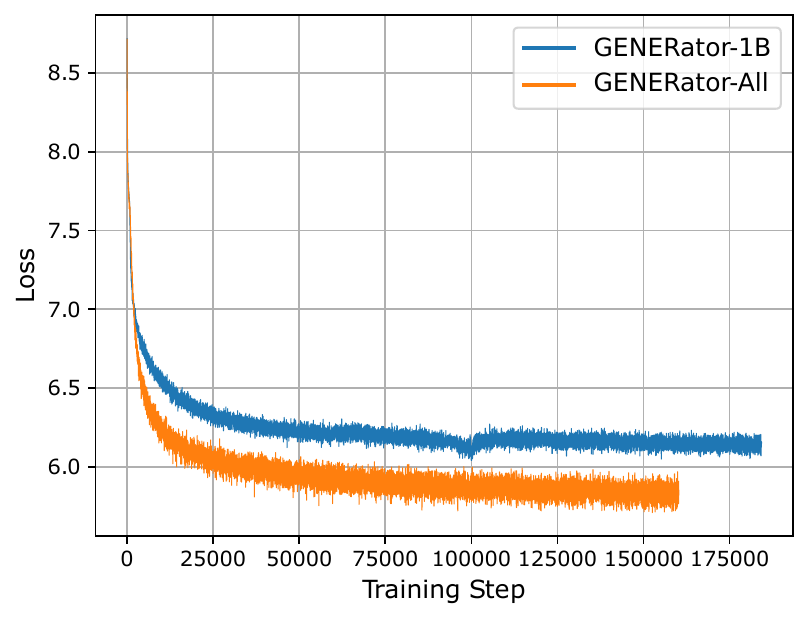}
    \caption{Comparison of pre-training loss between functional and all-sequence training paradigms. The plot shows pre-training loss curves for GENERator-1B (trained on functional sequences) and GENERator-All (trained on all genomic sequences). While GENERator-All achieves significantly lower pre-training loss, this comparison is misleading due to fundamental differences in training data composition. The all-sequence training paradigm includes extensive non-functional regions containing repetitive sequences and ambiguous segments (e.g., \texttt{NNNNNN}), which artificially depress the loss metric by introducing easily predictable patterns. In contrast, the functional sequence training strategy focuses exclusively on biologically active regions where DNA sequences exhibit inherently higher complexity and information density. The loss discrepancy thus reflects data characteristics rather than model capability, as evident by the suboptimal performance of GENERator-All in the sequence recovery task (Figure~\ref{fig:kmer_model}), which closely resembles the pre-training objective and uses the same test dataset. This explains the apparent paradox where GENERator-All achieves superior pre-training loss yet demonstrates inferior performance across downstream tasks.}
    \label{fig:pretrain_loss}
\end{figure}

\begin{figure}[ht]
    \centering
    \includegraphics[width=0.9\textwidth]{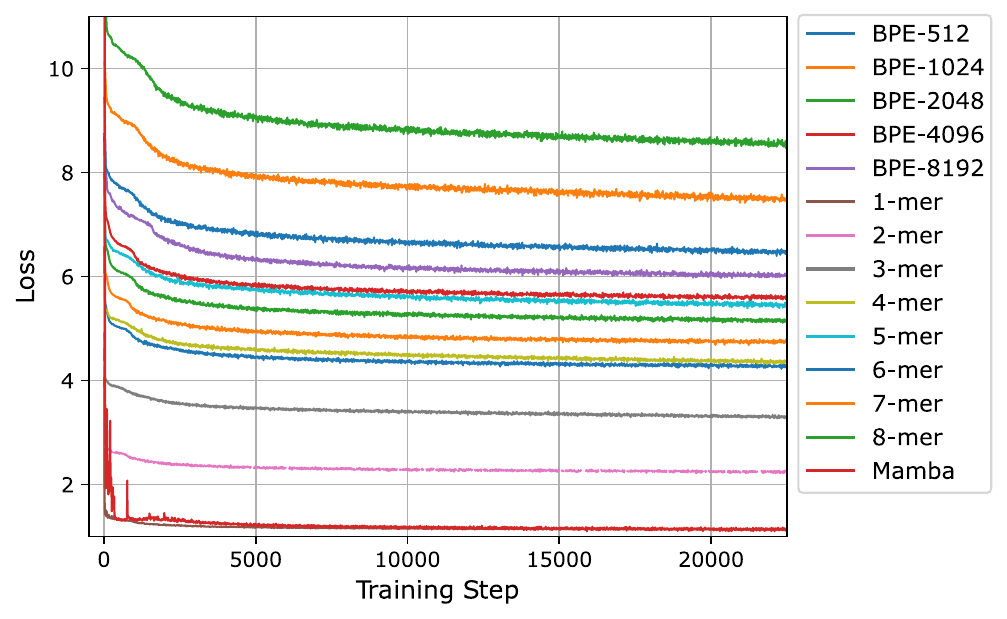}
    \caption{Comparison of pre-training loss across tokenization strategies. Loss curves are shown for various tokenizers: BPE (vocabulary sizes 512-8192), $k$-mer ($k$=1-8), and Mamba+1mer. The substantial variation in loss values—ranging from roughly 1 for 1-mer (4-way classification) to significantly higher for 8-mer (65536-way classification)—demonstrates the inherent incomparability of pre-training loss across different tokenization schemes. This fundamental limitation motivated the development of our sequence recovery task (Section~\ref{sec:kmer_predition}), which enables fair cross-tokenizer and cross-architecture comparison. Close examination reveals why BPE underperforms: in the BPE-512 vs. 4-mer comparison, 4-mer initially shows higher loss but narrows the gap with training; similarly, in BPE-4096 vs. 5-mer, 5-mer exhibits faster loss reduction over time. These patterns indicate that BPE tokenizers present greater learning challenges, supporting our hypothesis that next token prediction pre-training may not be optimally aligned with BPE tokenization for genomic sequences. This misalignment stems from the hierarchical nature of BPE vocabulary: unlike human language with clear lexical boundaries, DNA sequences lack explicit delimiters, creating nested relationships among BPE tokens. For example, when predicting target token \texttt{GCCT}, partially correct predictions like \texttt{G}, \texttt{GC}, or \texttt{GCC} are penalized as incorrect, unnecessarily complicating the training process. The Mamba+1mer model exhibits initial training instability, likely due to the extreme sequence length of 98k tokens, but gradually stabilizing and converging to a loss level comparable to 1-mer, consistent with its sequence recovery performance.}
    \label{fig:pretrain_loss_tokenizer}
\end{figure}

\paragraph{Data Statistics and Training Dynamics}
Table~\ref{tab:pretrain_data_statistics_species} and Table~\ref{tab:pretrain_data_statistics_gene} summarize pre-training data statistics in terms of nucleotide counts and gene distributions. Figure~\ref{fig:pretrain_loss} compares training losses between GENERator and GENERator-All. Although GENERator-All achieves lower pre-training loss, it underperforms on downstream tasks, likely due to inclusion of non-functional regions containing repetitive sequences that artificially reduce loss without improving functional understanding. Figure~\ref{fig:pretrain_loss_tokenizer} shows loss trajectories for different tokenizers, where vocabulary size variations complicate direct comparisons and motivate the sequence recovery metric for fair model evaluation.

\section{Benchmark Evaluation Details}

\subsection{Benchmark Datasets}
Our evaluation encompasses two established benchmarks—Nucleotide Transformer tasks~\cite{nucleotide-transformer} and Genomic Benchmarks~\cite{genomic-benchmarks}—alongside the proposed Gener tasks. The Nucleotide Transformer datasets are available in two versions: the \href{https://huggingface.co/datasets/InstaDeepAI/nucleotide_transformer_downstream_tasks}{\textcolor{blue}{original}} and \href{https://huggingface.co/datasets/InstaDeepAI/nucleotide_transformer_downstream_tasks_revised}{\textcolor{blue}{revised}} releases. Genomic Benchmarks data can be accessed \href{https://huggingface.co/katarinagresova}{\textcolor{blue}{here}}.

\subsection{Evaluation Metrics}
We adhere to the original evaluation protocols for each benchmark. Nucleotide Transformer tasks employ the Matthews correlation coefficient (MCC):
\begin{equation*}
\text{MCC} = \frac{\text{TP} \times \text{TN} - \text{FP} \times \text{FN}}{\sqrt{(\text{TP} + \text{FP})(\text{TP} + \text{FN})(\text{TN} + \text{FP})(\text{TN} + \text{FN})}},
\end{equation*}
where TP, TN, FP, and FN denote true positives, true negatives, false positives, and false negatives, respectively. Genomic Benchmarks utilize accuracy, while Gener tasks adopt the weighted F1 score for multi-class classification:
\begin{align*}
&\text{Precision}_i = \frac{\text{TP}_i}{\text{TP}_i + \text{FP}_i},&
\text{Recall}_i &= \frac{\text{TP}_i}{\text{TP}_i + \text{FN}_i}, \\
\text{F1}_i &= 2 \times \frac{\text{Precision}_i \times \text{Recall}_i}{\text{Precision}_i + \text{Recall}_i}, &
\text{F1}_{\text{weighted}} &= \sum_{i=1}^{n} w_i \times \text{F1}_i,
\end{align*}
where $w_i = n_i/N$ represents the weight for class $i$, $n_i$ is the sample count in class $i$, and $N$ is the total sample size.

\subsection{Baseline Models}
Our comparative analysis includes state-of-the-art models spanning diverse architectures, training data, and scales. Table~\ref{tab:baseline_introduction} summarizes key characteristics. Evaluation results for Nucleotide Transformer tasks are sourced from the original publication~\cite{nucleotide-transformer} where available; remaining baselines were evaluated consistently under our framework. Models were accessed via HuggingFace repositories:
\begin{itemize}
    \item Caduceus-Ph: \texttt{kuleshov-group/caduceus-ph\_seqlen-131k\_d\_model-256\_n\_layer-16}
    \item Caduceus-PS: \texttt{kuleshov-group/caduceus-ps\_seqlen-131k\_d\_model-256\_n\_layer-16}
    \item DNABERT-2: \texttt{zhihan1996/DNABERT-2-117M}
    \item Enformer: \texttt{EleutherAI/enformer-official-rough}
    \item Evo2-1B: \texttt{arcinstitute/evo2\_1b\_base}
    \item Evo2-7B: \texttt{arcinstitute/evo2\_7b\_base}
    \item GROVER: \texttt{PoetschLab/GROVER}
    \item HyenaDNA: \texttt{LongSafari/hyenadna-large-1m-seqlen}
    \item NT: \texttt{InstaDeepAI/nucleotide-transformer-2.5b-multi-species}
    \item NT-v2: \texttt{InstaDeepAI/nucleotide-transformer-v2-500m-multi-species}
\end{itemize}

\subsection{Experimental Configuration}
\label{sec:exp_config_supp}
To ensure fair comparison, all models were uniformly fine-tuned and evaluated using 10-fold cross-validation on each dataset. For every model-dataset combination, we conducted extensive hyperparameter searches, exploring learning rates in $\{1\times10^{-5}, 2\times10^{-5}, 5\times10^{-5}, 1\times10^{-4}, 2\times10^{-4}, 5\times10^{-4}, 1\times10^{-3}, 2\times10^{-3}, 5\times10^{-3}\}$ and batch sizes in $\{64, 128, 256, 512\}$. We maintained pre-training optimizer settings ($\beta_1=0.9$, $\beta_2=0.95$, weight decay $=0.1$) and implemented a reduce-on-plateau learning rate scheduler with early stopping (patience=5). 

Optimal hyperparameters identified through grid search are documented in Tables~\ref{tab:benchmark_hyperparam1}--\ref{tab:benchmark_hyperparam4}. Models operated within their context length constraints: full sequences for Nucleotide Transformer and Genomic Benchmarks, truncated inputs for longer Gener task sequences. For causal language models, predictions derived from the \texttt{<EOS>} token embedding via a linear layer; masked language models used \texttt{<BOS>} or \texttt{<CLS>} tokens. Caduceus employed mean pooling with reverse complement augmentation for Caduceus-Ph. All models utilized final-layer embeddings and underwent full fine-tuning. Metrics report 10-fold cross-validation averages.

\section{Central Dogma Implementation Details}
\label{sec:central_dogma_supp}

Protein sequences and corresponding coding DNA fragments for cytochrome P450 and histone families were sourced from UniProt~\cite{UniProt}, with approximately 400,000 entries per family. We fine-tuned GENERator on these protein-coding sequences using a learning rate of $5 \times 10^{-5}$ and batch size of 1024. For sequence generation, we produced 100,000 sequences per combination of temperature $T \in \{0.5, 0.6, 0.7, 0.8, 0.9, 1.0\}$ and nucleus sampling $P \in \{0.5, 0.6, 0.7, 0.8, 0.9, 1.0\}$. Following translation to protein sequences, duplicates were removed by comparison with training data, yielding 10,000 unique samples per family. Optimal hyperparameters were $T=0.6$, $P=0.6$ for cytochrome P450 and $T=1.0$, $P=0.8$ for histone.

\section{CRE Design Implementation Details}

\subsection{Model Training and Generation}
For CRE activity prediction, we followed DeepSTARR~\cite{DeepSTARR} data partitioning and employed the same hyperparameter search strategy as benchmark experiments. The activity predictor was trained with learning rate $2 \times 10^{-5}$ and batch size 64. CRE sequences from top and bottom activity quartiles in the DeepSTARR training set were used for supervised fine-tuning, labeled with \texttt{<high>} and \texttt{<low>} prompts respectively. SFT used learning rate $1 \times 10^{-5}$ and batch size 2048. Generation employed the same hyperparameter search strategy as Section~\ref{sec:central_dogma_supp}, with visualization parameters set to $T=0.5$, $P=0.7$ for developmental activity and $T=0.8$, $P=0.6$ for housekeeping activity in Figure~\ref{fig:enhancer_design}.

\subsection{Experimental Methods}

\paragraph{Construction of Plasmid screening Libraries and input libraries.}
Candidate oligonucleotides were synthesized and converted to double-stranded DNA (dsDNA) by PCR. PCR products were purified using a gel recovery kit (Zymo Research, Cat. D4008). The Drosophila STARR-seq plasmid containing DSCP/Rps12 core promoters was linearized by digestion with AgeI and SalI. Gibson assembly (New England Biolabs, Cat. E2611L) was used to ligate the dsDNA fragments into the linearized plasmid backbone. Recombinant plasmids were introduced into electrocompetent \emph{}{E. coli} by electroporation. Following electroporation, 5 mL of recovery medium was added, and cultures were incubated at 37 °C with shaking ($\sim$220 rpm) for 16 h. Plasmids were extracted from overnight cultures using a plasmid extraction kit (QIAGEN, Cat. 12362) and quantified for downstream applications. Input libraries were generated to quantify each sequence in the plasmid screening libraries. Experimental details were performed as previously described~\cite{neumayr2019starr}. Briefly, the screening plasmids were first amplified by PCR using gene-specific pUMI-containing primers. PCR products were purified using a DNA purification kit, amplified to incorporate adapters for next-generation sequencing, and sequenced by 300 bp paired-end sequencing (Zymo Research, Cat. D4014).

\paragraph{Cell Culture and Transfection}
S2 cells were maintained in 90\% Schneider’s Drosophila medium (Thermo Fisher Scientific, Cat. 21720024) supplemented with 10\% fetal bovine serum (Gibco, Cat. 10270-106) and 1\% penicillin--streptomycin (Gibco, Cat. 2441868) at 28 °C without CO\textsubscript{2}. For transfections, $1 \times 10^7$ cells and 20 µg plasmid DNA were used per reaction. Electroporation was performed with a 4D-Nucleofector\textregistered{} X Unit (Lonza, Cat. AAF-1003X) using program DS120.

\paragraph{Construction of output libraries}
Output libraries were generated to quantify the self-transcribed abundance of sequences under screening. Experimental details were performed as previously described~\cite{neumayr2019starr}. Briefly, total RNA was extracted from transfected cells 24 h post-transfection using an RNA extraction kit (MF167-01) according to the manufacturer’s protocol. mRNA was enriched and diluted to 100 µL with RNase-free water. Reverse transcription was performed at 60 °C for 5 min, followed by incubation at 55 °C for 30--60 min. For cDNA fragments $>$5 kb, 2 µL of amplification enzyme was added. Reactions using random primers were pre-incubated at 25 °C for 5 min. Enzymes were inactivated at 70 °C for 15 min. To remove RNA templates, RNase A and RNase H (1 µL each) were added and incubated at 37 °C for 1 h. The primers for the reverse transcription process are reverse transcription primers that contain UMI. cDNA was amplified by PCR using gene-specific primers in the first round and sequencing primers in the second round. Final PCR products were purified and sequenced using 300 bp paired-end sequencing.

\paragraph{Quality control and CRE activity quantification of UMI-STARR-Seq libraries}
Sequence abundance was quantified using UMI counts in both input and output libraries. Library complexity was assessed by plotting the distribution of UMI counts per sequence using the \texttt{hist()} function from the Python package \texttt{matplotlib.pyplot}. CRE activity was quantified with \texttt{DESeq2} and expressed as $\log_2$ fold changes between output and input UMI counts for each sequence. To assess reproducibility across biological replicates, CRE activity was compared between each pair of replicates. Comparisons were visualized as scatter plots using the \texttt{geom\_point()} function in the \texttt{ggplot2} package in R. Pearson correlation coefficients were calculated using the \texttt{stat\_poly\_eq()} function, and the corresponding regression lines were overlaid on the plots to illustrate the relationships.

\paragraph{Motif analysis and visualization}
Motif analysis was performed using the JASPAR2022 insect motif database. The \textit{Drosophila melanogaster} reference genome (\texttt{BSgenome.Dmelanogaster.UCSC.dm6}) was used as the background sequence. Motif occurrences in all sequences were predicted using the \texttt{matchMotifs()} function from the R package \texttt{motifmatchr}. Motif enrichment was evaluated using a hypergeometric test. For each group, the top ten enriched motifs were selected and visualized as dot plots using the \texttt{geom\_point()} function in \texttt{ggplot2}. For the sequences Read\_1602 and Read\_3778, base-level activity impact logos were generated using the Python package \texttt{logomaker}. Motif analysis of these two sequences was initially performed using the JASPAR insect motif database and was subsequently complemented by searches against the JASPAR vertebrate motif database.

\section{Computational Resources}
Model training and inference utilized NVIDIA V100 and A100 GPUs. Preliminary experiments involved training 14 models for one epoch to evaluate tokenizer performance. Pre-training encompassed two models with different training data. Downstream tasks required over 10,000 runs, including extensive hyperparameter search across 36 combinations and 10-fold cross-validation for each benchmark task per model. Detailed computational resource statistics are provided in Table~\ref{tab:computational_resources}.

\section{Supplementary Figures \& Tables}
\begin{figure}[!htb]
    \centering
    \includegraphics[width=\textwidth]{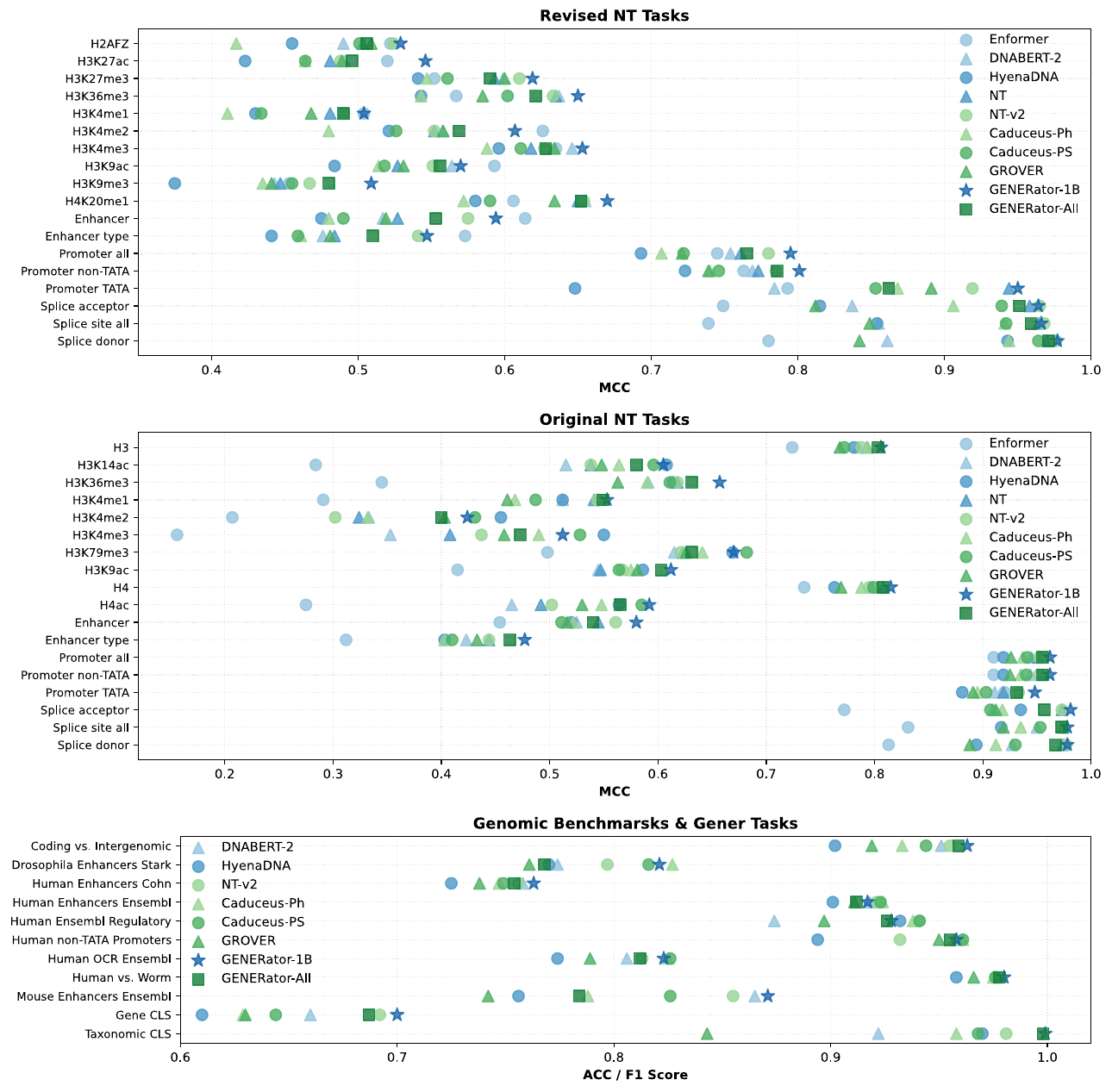}
    \caption{Detailed model performance across three benchmark suites. GENERator-1B achieves the highest average score across all tasks while demonstrating exceptional performance in individual benchmarks, securing first place in 32 out of 47 tasks and second place in 10 tasks. It consistently outperforms the control model GENERator-All, validating the effectiveness of functional sequence training.}
    \label{fig:benchmarks_supp}
\end{figure}
\begin{figure}[!htb]
    \centering
    \includegraphics[width=0.95\textwidth]{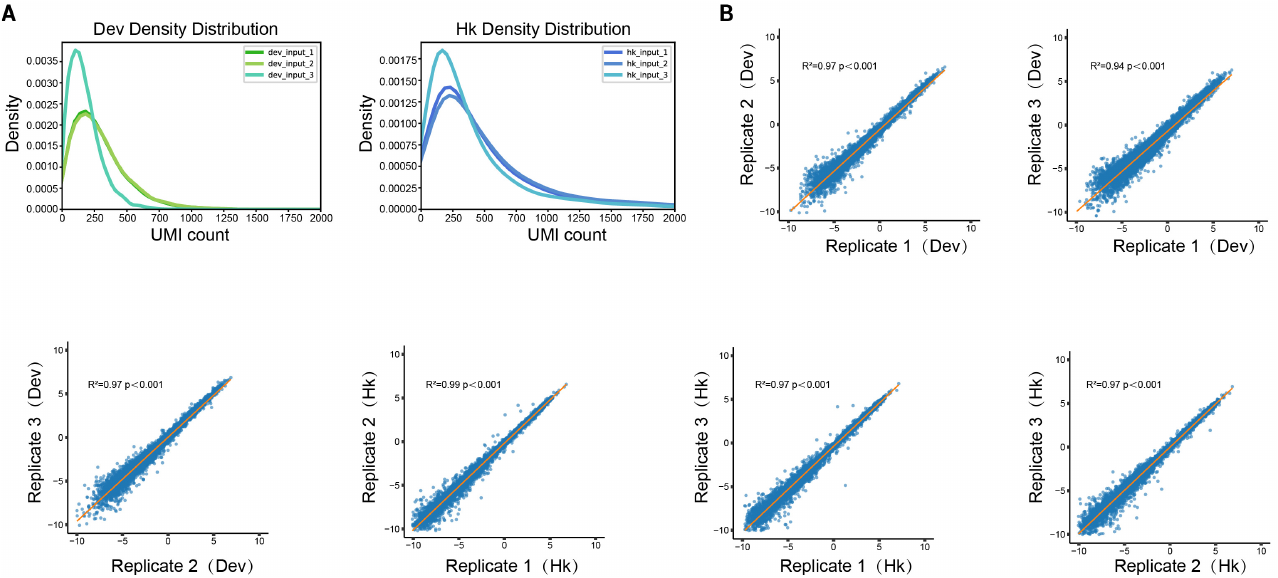}
    \caption{Quality control of UMI-STARR-Seq. (A) Density distribution of UMI counts per sequence in the input libraries. (B) Correlation of CRE activity ($\log_2$ fold change of output over input) across biological replicates of UMI-STARR-Seq.}
    \label{fig:enhancer_supp_1}
\end{figure}

\begin{figure}[!htb]
    \centering
    \includegraphics[width=0.95\textwidth]{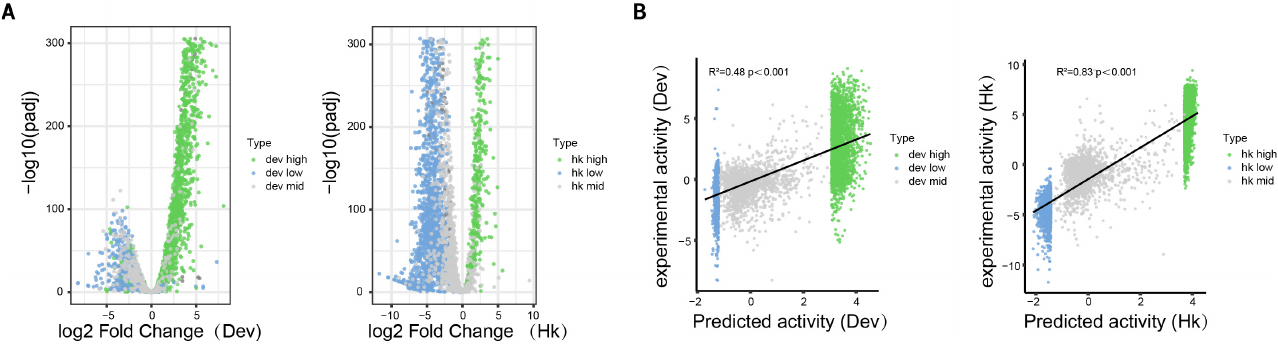}
    \caption{Experimental activity of GENERator-designed CREs. (A) Volcano plots of CRE activity, colored by predicted activity categories. (B) Correlation between experimentally measured activity and predicted activity for all GENERator-designed sequences. Both the x- and y-axes represent $\log_2$-transformed CRE activity.}
    \label{fig:enhancer_supp_2}
\end{figure}

\begin{figure}[!htb]
    \centering
    \includegraphics[width=0.95\textwidth]{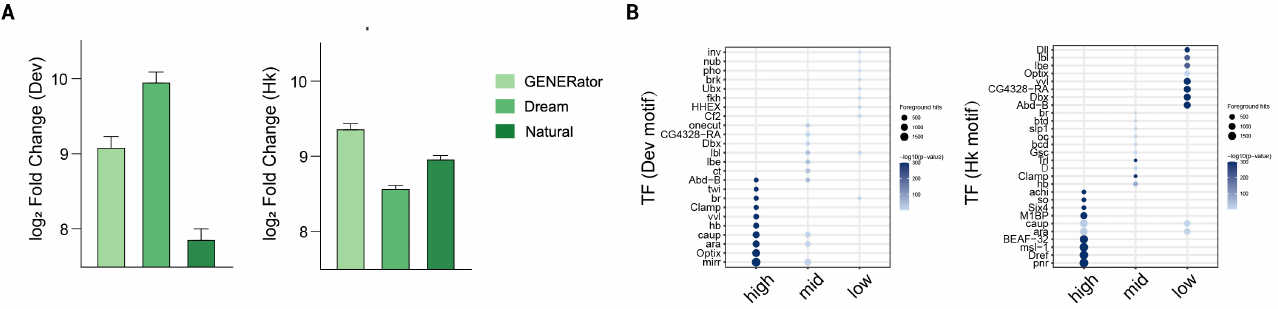}
    \caption{Experimental measurement and transcription factor motif enrichment of GENERator-designed CREs. (A) Comparison of the activities of the highest-activity sequences from GENERator-designed, DREAM-designed, and natural CREs. (B) Enrichment of transcription factor DNA-binding motifs in GENERator-designed sequences, stratified by predicted high, medium, and low activity categories.}
    \label{fig:enhancer_supp_3}
\end{figure}

\clearpage

\begin{table*}[!htb]
\small
\renewcommand{\arraystretch}{1.2}
\centering
\caption{Zero-shot test of sequence recovery. The metric represents the accuracy of next 30 bp prediction, measuring the overlap between the first 30 bp of generated subsequent sequences and the reference sequence. Models were evaluated with input sequences of 6144 bp or the maximum context length supported by each model.}
\resizebox{\textwidth}{!}{
\begin{tabular}{lccccccccc}
\toprule
 & \multicolumn{4}{c}{\textbf{Masked Language Models}} & \multicolumn{5}{c}{\textbf{Causal Language Models}} \\
\cmidrule(lr){2-5} \cmidrule(lr){6-10}
 & DNABERT-2 & GROVER & NT & NT-v2 & HyenaDNA & Evo2-1B & Evo2-7B & GENERator-1B & GENERator-3B \\
\midrule
Protozoa     & 0.289 & \textbf{0.305} & 0.301 & \underline{0.304} & 0.315 & \underline{0.406} & \textbf{0.429} & 0.382 & 0.392 \\
Fungi        & 0.266 & 0.280 & \underline{0.305} & \textbf{0.314} & 0.282 & 0.442 & \textbf{0.521} & 0.428 & \underline{0.478} \\
Plant        & 0.272 & 0.282 & \underline{0.316} & \textbf{0.321} & 0.297 & 0.517 & \textbf{0.616} & 0.533 & \underline{0.584} \\
Invertebrate & 0.274 & 0.284 & \underline{0.345} & \textbf{0.351} & 0.300 & 0.451 & \textbf{0.521} & 0.451 & \underline{0.488} \\
Mammalian    & 0.280 & \underline{0.291} & \textbf{0.325} & \textbf{0.325} & 0.297 & 0.633 & \textbf{0.787} & 0.684 & \underline{0.751} \\
Vertebrate Other & 0.275 & 0.284 & \underline{0.334} & \textbf{0.342} & 0.293 & 0.578 & \textbf{0.697} & 0.613 & \underline{0.676} \\
Overall      & 0.276 & 0.288 & \underline{0.321} & \textbf{0.326} & 0.297 & 0.504 & \textbf{0.595} & 0.515 & \underline{0.561} \\
\bottomrule
\end{tabular}
}
\label{tab:sequence_recovery_supp}
\end{table*}
\begin{table*}[!htb]
\small
\renewcommand{\arraystretch}{1.2}
\centering
\caption{Zero-shot test of variant effect prediction on ClinVar dataset.}
\resizebox{\textwidth}{!}{
\begin{tabular}{lcccccccccccc}
\toprule
 & \multicolumn{5}{c}{\textbf{MSA Required}} & \multicolumn{7}{c}{\textbf{Sequence Only}} \\
\cmidrule(lr){2-6} \cmidrule(lr){7-13}
 & phastCons-100v & phyloP-100v & phyloP-241m & CADD & GPN-MSA & NT & NT-v2 & HyenaDNA & Evo2-1B & Evo2-7B & GENERator-1B & GENERator-3B \\
\midrule
AUROC & 0.883 & 0.927 & 0.912 & \underline{0.966} & \textbf{0.970} & 0.580 & 0.665 & 0.503 & 0.867 & \textbf{0.921} & \underline{0.910} & \textbf{0.921} \\
AUPRC & 0.848 & 0.937 & 0.913 & \underline{0.967} & \textbf{0.974} & 0.630 & 0.733 & 0.546 & 0.906 & \textbf{0.942} & 0.930 & \underline{0.939} \\
\bottomrule
\end{tabular}
}
\label{tab:variant_effect_prediction}
\end{table*}
\begin{table*}[!htb]
\small
\renewcommand{\arraystretch}{1.2}
\centering
\caption{Evaluation of the revised Nucleotide Transformer tasks. The reported values represent the Matthews correlation coefficient (MCC) averaged over 10-fold cross-validation, with the standard error in parentheses.}
\resizebox{\textwidth}{!}{
\begin{tabular}{lcccccccccc}
\toprule
& Enformer & DNABERT-2 & HyenaDNA & NT & NT-v2 & Caduceus-Ph & Caduceus-PS & GROVER & GENERator & GENERator-All \\
& (252M) & (117M) & (55M) & (2.5B) & (500M) & (8M) & (8M) & (87M) & (1.2B) & (1.2B) \\
\midrule
H2AFZ          & 0.522 (0.019) & 0.490 (0.013) & 0.455 (0.015) & 0.503 (0.010) & \underline{0.524 (0.008)} & 0.417 (0.016) & 0.501 (0.013) & 0.509 (0.013) & \textbf{0.529 (0.009)} & 0.506 (0.019) \\
H3K27ac        & \underline{0.520 (0.015)} & 0.491 (0.010) & 0.423 (0.017) & 0.481 (0.020) & 0.488 (0.013) & 0.464 (0.018) & 0.464 (0.022) & 0.489 (0.023) & \textbf{0.546 (0.015)} & 0.496 (0.014) \\
H3K27me3       & 0.552 (0.007) & 0.599 (0.010) & 0.541 (0.018) & 0.593 (0.016) & \underline{0.610 (0.006)} & 0.547 (0.010) & 0.561 (0.036) & 0.600 (0.008) & \textbf{0.619 (0.008)} & 0.590 (0.014) \\
H3K36me3       & 0.567 (0.017) & \underline{0.637 (0.007)} & 0.543 (0.010) & 0.635 (0.016) & 0.633 (0.015) & 0.543 (0.009) & 0.602 (0.008) & 0.585 (0.008) & \textbf{0.650 (0.006)} & 0.621 (0.013) \\
H3K4me1        & \textbf{0.504 (0.021)} & \underline{0.490 (0.008)} & 0.430 (0.014) & 0.481 (0.012) & \underline{0.490 (0.017)} & 0.411 (0.012) & 0.434 (0.030) & 0.468 (0.011) & \textbf{0.504 (0.010)} & \underline{0.490 (0.016)} \\
H3K4me2        & \textbf{0.626 (0.015)} & 0.558 (0.013) & 0.521 (0.024) & 0.552 (0.022) & 0.552 (0.013) & 0.480 (0.013) & 0.526 (0.035) & 0.558 (0.012) & \underline{0.607 (0.010)} & 0.569 (0.012) \\
H3K4me3        & 0.635 (0.019) & \underline{0.646 (0.008)} & 0.596 (0.015) & 0.618 (0.015) & 0.627 (0.020) & 0.588 (0.020) & 0.611 (0.015) & 0.634 (0.011) & \textbf{0.653 (0.008)} & 0.628 (0.018) \\
H3K9ac         & \textbf{0.593 (0.020)} & 0.564 (0.013) & 0.484 (0.022) & 0.527 (0.017) & 0.551 (0.016) & 0.514 (0.014) & 0.518 (0.018) & 0.531 (0.014) & \underline{0.570 (0.017)} & 0.556 (0.018) \\
H3K9me3        & 0.453 (0.016) & 0.443 (0.025) & 0.375 (0.026) & 0.447 (0.018) & 0.467 (0.044) & 0.435 (0.019) & 0.455 (0.019) & 0.441 (0.017) & \textbf{0.509 (0.013)} & \underline{0.480 (0.037)} \\
H4K20me1       & 0.606 (0.016) & \underline{0.655 (0.011)} & 0.580 (0.009) & 0.650 (0.014) & 0.654 (0.011) & 0.572 (0.012) & 0.590 (0.020) & 0.634 (0.006) & \textbf{0.670 (0.006)} & 0.652 (0.010) \\
Enhancer       & \textbf{0.614 (0.010)} & 0.517 (0.011) & 0.475 (0.006) & 0.527 (0.012) & 0.575 (0.023) & 0.480 (0.008) & 0.490 (0.009) & 0.519 (0.009) & \underline{0.594 (0.013)} & 0.553 (0.020) \\
Enhancer type & \textbf{0.573 (0.013)} & 0.476 (0.009) & 0.441 (0.010) & 0.484 (0.012) & 0.541 (0.013) & 0.461 (0.009) & 0.459 (0.011) & 0.481 (0.009) & \underline{0.547 (0.017)} & 0.510 (0.022) \\
Promoter all   & 0.745 (0.012) & 0.754 (0.009) & 0.693 (0.016) & 0.761 (0.009) & \underline{0.780 (0.012)} & 0.707 (0.017) & 0.722 (0.014) & 0.721 (0.011) & \textbf{0.795 (0.005)} & 0.765 (0.009) \\
Promoter non-TATA & 0.763 (0.012) & 0.769 (0.009) & 0.723 (0.013) & 0.773 (0.010) & 0.785 (0.009) & 0.740 (0.012) & 0.746 (0.009) & 0.739 (0.018) & \textbf{0.801 (0.005)} & \underline{0.786 (0.007)} \\
Promoter TATA  & 0.793 (0.026) & 0.784 (0.036) & 0.648 (0.044) & \underline{0.944 (0.016)} & 0.919 (0.028) & 0.868 (0.023) & 0.853 (0.034) & 0.891 (0.041) & \textbf{0.950 (0.009)} & 0.862 (0.024) \\
Splice acceptor & 0.749 (0.007) & 0.837 (0.006) & 0.815 (0.049) & 0.958 (0.003) & \textbf{0.965 (0.004)} & 0.906 (0.015) & 0.939 (0.012) & 0.812 (0.012) & \underline{0.964 (0.003)} & 0.951 (0.006) \\
Splice site all & 0.739 (0.011) & 0.855 (0.005) & 0.854 (0.053) & 0.964 (0.003) & \textbf{0.968 (0.003)} & 0.941 (0.006) & 0.942 (0.012) & 0.849 (0.015) & \underline{0.966 (0.003)} & 0.959 (0.003) \\
Splice donor   & 0.780 (0.007) & 0.861 (0.004) & 0.943 (0.024) & 0.970 (0.002) & \underline{0.976 (0.003)} & 0.944 (0.026) & 0.964 (0.010) & 0.842 (0.009) & \textbf{0.977 (0.002)} & 0.971 (0.002) \\
\bottomrule
\end{tabular}
}
\label{tab:nucleotide_transformer_tasks_revised}
\end{table*}
\begin{table*}[!htb]
\small
\renewcommand{\arraystretch}{1.2}
\centering
\caption{Evaluation of the original Nucleotide Transformer tasks. The reported values represent the Matthews correlation coefficient (MCC) averaged over 10-fold cross-validation, with the standard error in parentheses.}
\resizebox{\textwidth}{!}{%
\begin{tabular}{lcccccccccc}
\toprule
& Enformer & DNABERT-2 & HyenaDNA & NT & NT-v2 & Caduceus-Ph & Caduceus-PS & GROVER & GENERator & GENERator-All \\
& (252M) & (117M) & (55M) & (2.5B) & (500M) & (8M) & (8M) & (87M) & (1.2B) & (1.2B) \\
\midrule
H3 & 0.724 (0.018) & 0.785 (0.012) & 0.781 (0.015) & 0.793 (0.013) & 0.788 (0.010) & 0.794 (0.012) & 0.772 (0.022) & 0.768 (0.008) & \textbf{0.806 (0.005)} & \underline{0.803 (0.007)} \\
H3K14ac & 0.284 (0.024) & 0.515 (0.009) & \textbf{0.608 (0.020)} & 0.538 (0.009) & 0.538 (0.015) & 0.564 (0.033) & 0.596 (0.038) & 0.548 (0.020) & \underline{0.605 (0.008)} & 0.580 (0.038) \\
H3K36me3 & 0.345 (0.019) & 0.591 (0.005) & 0.614 (0.014) & 0.618 (0.011) & 0.618 (0.015) & 0.590 (0.018) & 0.611 (0.048) & 0.563 (0.017) & \textbf{0.657 (0.007)} & \underline{0.631 (0.013)} \\
H3K4me1 & 0.291 (0.016) & 0.512 (0.008) & 0.512 (0.008) & 0.541 (0.005) & 0.544 (0.009) & 0.468 (0.015) & 0.487 (0.029) & 0.461 (0.018) & \textbf{0.553 (0.009)} & \underline{0.549 (0.018)} \\
H3K4me2 & 0.207 (0.021) & 0.333 (0.013) & \textbf{0.455 (0.028)} & 0.324 (0.014) & 0.302 (0.020) & 0.332 (0.034) & \underline{0.431 (0.016)} & 0.403 (0.042) & 0.424 (0.013) & 0.400 (0.015) \\
H3K4me3 & 0.156 (0.022) & 0.353 (0.021) & \textbf{0.550 (0.015)} & 0.408 (0.011) & 0.437 (0.028) & 0.490 (0.042) & \underline{0.528 (0.033)} & 0.458 (0.022) & 0.512 (0.009) & 0.473 (0.047) \\
H3K79me3 & 0.498 (0.013) & 0.615 (0.010) & 0.669 (0.014) & 0.623 (0.010) & 0.621 (0.012) & 0.641 (0.028) & \textbf{0.682 (0.018)} & 0.626 (0.026) & \underline{0.670 (0.011)} & 0.631 (0.021) \\
H3K9ac & 0.415 (0.020) & 0.545 (0.009) & 0.586 (0.021) & 0.547 (0.011) & 0.567 (0.020) & 0.575 (0.024) & 0.564 (0.018) & 0.581 (0.015) & \textbf{0.612 (0.006)} & \underline{0.603 (0.019)} \\
H4 & 0.735 (0.023) & 0.797 (0.008) & 0.763 (0.012) & \underline{0.808 (0.007)} & 0.795 (0.008) & 0.788 (0.010) & 0.799 (0.010) & 0.769 (0.017) & \textbf{0.815 (0.008)} & \underline{0.808 (0.010)} \\
H4ac & 0.275 (0.022) & 0.465 (0.013) & 0.564 (0.011) & 0.492 (0.014) & 0.502 (0.025) & 0.548 (0.027) & \underline{0.585 (0.018)} & 0.530 (0.017) & \textbf{0.592 (0.015)} & 0.565 (0.035) \\
Enhancer & 0.454 (0.029) & 0.525 (0.026) & 0.520 (0.031) & 0.545 (0.028) & \underline{0.561 (0.029)} & 0.522 (0.024) & 0.511 (0.026) & 0.516 (0.018) & \textbf{0.580 (0.015)} & 0.540 (0.026) \\
Enhancer type & 0.312 (0.043) & 0.423 (0.018) & 0.403 (0.056) & 0.444 (0.022) & 0.444 (0.036) & 0.403 (0.028) & 0.410 (0.026) & 0.433 (0.029) & \textbf{0.477 (0.017)} & \underline{0.463 (0.023)} \\
Promoter all & 0.910 (0.004) & 0.945 (0.003) & 0.919 (0.003) & 0.951 (0.004) & 0.952 (0.002) & 0.937 (0.002) & 0.941 (0.003) & 0.926 (0.004) & \textbf{0.962 (0.002)} & \underline{0.955 (0.002)} \\
Promoter non-TATA & 0.910 (0.006) & 0.944 (0.003) & 0.919 (0.004) & \underline{0.955 (0.003)} & 0.952 (0.003) & 0.935 (0.007) & 0.940 (0.002) & 0.925 (0.006) & \textbf{0.962 (0.001)} & \underline{0.955 (0.002)} \\
Promoter TATA & 0.920 (0.012) & 0.911 (0.011) & 0.881 (0.020) & 0.919 (0.008) & \underline{0.933 (0.009)} & 0.895 (0.010) & 0.903 (0.010) & 0.891 (0.009) & \textbf{0.948 (0.008)} & 0.931 (0.007) \\
Splice acceptor & 0.772 (0.007) & 0.909 (0.004) & 0.935 (0.005) & \underline{0.973 (0.002)} & \underline{0.973 (0.004)} & 0.918 (0.017) & 0.907 (0.015) & 0.912 (0.010) & \textbf{0.981 (0.002)} & 0.957 (0.009) \\
Splice site all & 0.831 (0.012) & 0.950 (0.003) & 0.917 (0.006) & 0.974 (0.004) & \underline{0.975 (0.002)} & 0.935 (0.011) & 0.953 (0.005) & 0.919 (0.005) & \textbf{0.978 (0.001)} & 0.973 (0.002) \\
Splice donor & 0.813 (0.015) & 0.927 (0.003) & 0.894 (0.013) & 0.974 (0.002) & \underline{0.977 (0.007)} & 0.912 (0.009) & 0.930 (0.010) & 0.888 (0.012) & \textbf{0.978 (0.002)} & 0.967 (0.005) \\
\bottomrule
\end{tabular}
}
\label{tab:nucleotide_transformer_tasks}
\end{table*}
\begin{table*}[!htb]
\small
\renewcommand{\arraystretch}{1.2}
\centering
\caption{Evaluation of the Genomic Benchmarks. The reported values represent the accuracy averaged over 10-fold cross-validation, with the standard error in parentheses.}
\resizebox{\textwidth}{!}{
\begin{tabular}{lcccccccc}
\toprule
& DNABERT-2 & HyenaDNA & NT-v2 & Caduceus-Ph & Caduceus-PS & GROVER & GENERator & GENERator-All \\
& (117M) & (55M) & (500M) & (8M) & (8M) & (87M) & (1.2B) & (1.2B) \\
\midrule
Coding vs. Intergenomic & 0.951 (0.002) & 0.902 (0.004) & 0.955 (0.001) & 0.933 (0.001) & 0.944 (0.002) & 0.919 (0.002) & \textbf{0.963 (0.000)} & \underline{0.959 (0.001)} \\
Drosophila Enhancers Stark & 0.774 (0.011) & 0.770 (0.016) & 0.797 (0.009) & \textbf{0.827 (0.010)} & 0.816 (0.015) & 0.761 (0.011) & \underline{0.821 (0.005)} & 0.768 (0.015) \\
Human Enhancers Cohn & \underline{0.758 (0.005)} & 0.725 (0.009) & 0.756 (0.006) & 0.747 (0.003) & 0.749 (0.003) & 0.738 (0.003) & \textbf{0.763 (0.002)} & 0.754 (0.006) \\
Human Enhancers Ensembl & 0.918 (0.003) & 0.901 (0.003) & 0.921 (0.004) & \textbf{0.924 (0.002)} & \underline{0.923 (0.002)} & 0.911 (0.004) & 0.917 (0.002) & 0.912 (0.002) \\
Human Ensembl Regulatory & 0.874 (0.007) & 0.932 (0.001) & \textbf{0.941 (0.001)} & \underline{0.938 (0.004)} & \textbf{0.941 (0.002)} & 0.897 (0.001) & 0.928 (0.001) & 0.926 (0.001) \\
Human non-TATA Promoters & 0.957 (0.008) & 0.894 (0.023) & 0.932 (0.006) & \textbf{0.961 (0.003)} & \textbf{0.961 (0.002)} & 0.950 (0.005) & \underline{0.958 (0.001)} & 0.955 (0.005) \\
Human OCR Ensembl & 0.806 (0.003) & 0.774 (0.004) & 0.813 (0.001) & \underline{0.825 (0.004)} & \textbf{0.826 (0.003)} & 0.789 (0.002) & 0.823 (0.002) & 0.812 (0.003) \\
Human vs. Worm & 0.977 (0.001) & 0.958 (0.004) & 0.976 (0.001) & 0.975 (0.001) & 0.976 (0.001) & 0.966 (0.001) & \textbf{0.980 (0.000)} & \underline{0.978 (0.001)} \\
Mouse Enhancers Ensembl & \underline{0.865 (0.014)} & 0.756 (0.030) & 0.855 (0.018) & 0.788 (0.028) & 0.826 (0.021) & 0.742 (0.025) & \textbf{0.871 (0.015)} & 0.784 (0.027) \\
\bottomrule
\end{tabular}
}
\label{tab:genomic_benchmarks}
\end{table*}
\begin{table*}[!htb]
\small
\renewcommand{\arraystretch}{1.2}
\centering
\caption{Evaluation of the Gener tasks. The reported values represent the weighted F1 score averaged over 10-fold cross-validation, with the standard error in parentheses.}
\resizebox{\textwidth}{!}{
\begin{tabular}{lcccccccc}
\toprule
& DNABERT-2 & HyenaDNA & NT-v2 & Caduceus-Ph & Caduceus-PS & GROVER & GENERator & GENERator-All \\
& (117M) & (55M) & (500M) & (8M) & (8M) & (87M) & (1.2B) & (1.2B) \\
\midrule
Gene CLS & 0.660 (0.002) & 0.610 (0.007) & \underline{0.692 (0.005)} & 0.629 (0.005) & 0.644 (0.007) & 0.630 (0.003) & \textbf{0.700 (0.002)} & 0.687 (0.003) \\
Taxonomic CLS & 0.922 (0.003) & 0.970 (0.024) & 0.981 (0.001) & 0.958 (0.021) & 0.968 (0.006) & 0.843 (0.006) & \textbf{0.999 (0.000)} & \underline{0.998 (0.001)} \\
\bottomrule
\end{tabular}
}
\label{tab:gener_tasks}
\end{table*}
\begin{table}[!htb]
\small
\renewcommand{\arraystretch}{1.2}
\centering
\caption{Evaluation of the CRE activity prediction on DeepSTARR hold-out test set. The reported values represent the Pearson correlation coefficient.}
\begin{tabular}{lcccc}
\toprule
 & DeepSTARR & NT & GENERator\\
\midrule
Dev & \underline{0.68} & 0.64 & \textbf{0.71} \\
Hk & 0.74 & \underline{0.75} & \textbf{0.80} \\
\bottomrule
\end{tabular}
\label{tab:enhancer_benchmark}
\end{table}
\begin{table*}[!htb]
\small
\renewcommand{\arraystretch}{0.8}
\centering
\caption{Hyperparameter settings for the original Nucleotide Transformer tasks.}
\begin{tabular}{@{}l*{20}{c}@{}}
\toprule
 & 
\multicolumn{2}{c}{\shortstack{NT-v2}} & 
\multicolumn{2}{c}{\shortstack{Caduceus-Ph}} & 
\multicolumn{2}{c}{\shortstack{Caduceus-PS}} & 
\multicolumn{2}{c}{\shortstack{GROVER}} & 
\multicolumn{2}{c}{\shortstack{GENERator}} & 
\multicolumn{2}{c}{\shortstack{GENERator-All}} \\ 
\cmidrule(l){2-3} 
\cmidrule(l){4-5} 
\cmidrule(l){6-7} 
\cmidrule(l){8-9} 
\cmidrule(l){10-11} 
\cmidrule(l){12-13}
 & LR & BS & LR & BS & LR & BS & LR & BS & LR & BS & LR & BS \\ 
\midrule
H3 & $2e^{-5}$ & 256 & $5e^{-4}$ & 64 & $1e^{-3}$ & 128 & $1e^{-4}$ & 256 & $1e^{-4}$ & 64 & $2e^{-5}$ & 64 \\
H3K14ac & $1e^{-5}$ & 64 & $2e^{-4}$ & 64 & $5e^{-4}$ & 64 & $5e^{-4}$ & 512 & $2e^{-4}$ & 512 & $5e^{-5}$ & 128 \\
H3K36me3 & $2e^{-5}$ & 64 & $5e^{-4}$ & 128 & $1e^{-3}$ & 256 & $5e^{-5}$ & 64 & $2e^{-4}$ & 512 & $1e^{-5}$ & 64 \\
H3K4me1 & $5e^{-5}$ & 64 & $2e^{-4}$ & 64 & $1e^{-3}$ & 512 & $5e^{-5}$ & 128 & $1e^{-4}$ & 128 & $2e^{-5}$ & 256 \\
H3K4me2 & $5e^{-5}$ & 256 & $2e^{-4}$ & 128 & $5e^{-4}$ & 512 & $1e^{-4}$ & 256 & $5e^{-5}$ & 256 & $1e^{-4}$ & 512 \\
H3K4me3 & $5e^{-5}$ & 128 & $5e^{-4}$ & 256 & $5e^{-4}$ & 256 & $1e^{-4}$ & 128 & $5e^{-5}$ & 64 & $5e^{-5}$ & 64 \\
H3K79me3 & $5e^{-5}$ & 128 & $2e^{-4}$ & 64 & $5e^{-4}$ & 256 & $1e^{-4}$ & 512 & $2e^{-4}$ & 128 & $5e^{-5}$ & 128 \\
H3K9ac & $5e^{-5}$ & 256 & $5e^{-4}$ & 256 & $5e^{-4}$ & 64 & $1e^{-4}$ & 64 & $1e^{-4}$ & 256 & $1e^{-4}$ & 256 \\
H4 & $2e^{-5}$ & 128 & $1e^{-4}$ & 128 & $5e^{-4}$ & 128 & $5e^{-5}$ & 512 & $5e^{-5}$ & 512 & $2e^{-5}$ & 128 \\
H4ac & $5e^{-5}$ & 64 & $5e^{-4}$ & 256 & $2e^{-4}$ & 64 & $5e^{-5}$ & 64 & $1e^{-4}$ & 128 & $5e^{-5}$ & 64 \\
Enhancers & $2e^{-5}$ & 128 & $2e^{-4}$ & 128 & $5e^{-4}$ & 512 & $5e^{-5}$ & 512 & $1e^{-4}$ & 512 & $2e^{-5}$ & 128 \\
Enhancers types & $1e^{-4}$ & 128 & $2e^{-4}$ & 256 & $2e^{-4}$ & 64 & $1e^{-4}$ & 512 & $2e^{-4}$ & 512 & $1e^{-4}$ & 128 \\
Promoter all & $1e^{-5}$ & 64 & $1e^{-4}$ & 64 & $2e^{-4}$ & 64 & $1e^{-4}$ & 64 & $1e^{-5}$ & 64 & $1e^{-5}$ & 64 \\
Promoter non-TATA & $5e^{-5}$ & 256 & $1e^{-3}$ & 512 & $5e^{-4}$ & 512 & $1e^{-4}$ & 256 & $2e^{-5}$ & 64 & $5e^{-5}$ & 256 \\
Promoter TATA & $5e^{-5}$ & 128 & $2e^{-3}$ & 512 & $1e^{-3}$ & 256 & $2e^{-4}$ & 512 & $5e^{-5}$ & 512 & $5e^{-5}$ & 128 \\
Splice sites acceptors & $2e^{-5}$ & 64 & $1e^{-3}$ & 64 & $2e^{-3}$ & 512 & $1e^{-4}$ & 64 & $2e^{-5}$ & 128 & $5e^{-5}$ & 64 \\
Splice sites all & $5e^{-5}$ & 128 & $2e^{-3}$ & 256 & $1e^{-3}$ & 64 & $1e^{-4}$ & 64 & $2e^{-5}$ & 256 & $5e^{-5}$ & 128 \\
Splice sites donors & $5e^{-5}$ & 128 & $1e^{-3}$ & 64 & $1e^{-3}$ & 64 & $5e^{-5}$ & 64 & $1e^{-4}$ & 64 & $5e^{-5}$ & 128 \\ 
\bottomrule
\end{tabular}
\label{tab:benchmark_hyperparam1}
\end{table*}

\begin{table*}[!htb]
\small
\renewcommand{\arraystretch}{0.8}
\centering
\caption{Hyperparameter settings for the revised Nucleotide Transformer tasks.}
\begin{tabular}{@{}l*{12}{c}@{}}
\toprule
 & 
\multicolumn{2}{c}{\shortstack{NT-v2}} & 
\multicolumn{2}{c}{\shortstack{Caduceus-Ph}} & 
\multicolumn{2}{c}{\shortstack{Caduceus-PS}} & 
\multicolumn{2}{c}{\shortstack{GROVER}} & 
\multicolumn{2}{c}{\shortstack{GENERator}} & 
\multicolumn{2}{c}{\shortstack{GENERator-All}} \\ 
\cmidrule(l){2-3} 
\cmidrule(l){4-5} 
\cmidrule(l){6-7} 
\cmidrule(l){8-9} 
\cmidrule(l){10-11} 
\cmidrule(l){12-13}
 & LR & BS & LR & BS & LR & BS & LR & BS & LR & BS & LR & BS \\ 
\midrule
H2AFZ & $2e^{-5}$ & 64 & $1e^{-5}$ & 64 & $1e^{-3}$ & 256 & $1e^{-5}$ & 128 & $2e^{-5}$ & 128 & $1e^{-4}$ & 128 \\
H3K27ac & $2e^{-5}$ & 128 & $5e^{-4}$ & 64 & $5e^{-4}$ & 64 & $2e^{-5}$ & 128 & $2e^{-5}$ & 256 & $2e^{-5}$ & 128 \\
H3K27me3 & $5e^{-5}$ & 256 & $1e^{-3}$ & 256 & $1e^{-3}$ & 128 & $1e^{-5}$ & 64 & $5e^{-5}$ & 256 & $5e^{-5}$ & 256 \\
H3K36me3 & $5e^{-5}$ & 128 & $2e^{-4}$ & 64 & $1e^{-3}$ & 256 & $5e^{-5}$ & 256 & $1e^{-5}$ & 256 & $5e^{-5}$ & 128 \\
H3K4me1 & $5e^{-5}$ & 512 & $5e^{-4}$ & 256 & $1e^{-3}$ & 256 & $5e^{-5}$ & 128 & $5e^{-5}$ & 64 & $1e^{-5}$ & 64 \\
H3K4me2 & $2e^{-5}$ & 256 & $1e^{-3}$ & 256 & $1e^{-3}$ & 256 & $2e^{-5}$ & 128 & $2e^{-5}$ & 64 & $5e^{-5}$ & 512 \\
H3K4me3 & $2e^{-5}$ & 64 & $1e^{-3}$ & 128 & $5e^{-4}$ & 64 & $1e^{-5}$ & 128 & $2e^{-5}$ & 64 & $5e^{-5}$ & 64 \\
H3K9ac & $5e^{-5}$ & 512 & $5e^{-4}$ & 128 & $1e^{-4}$ & 64 & $1e^{-5}$ & 64 & $2e^{-5}$ & 128 & $5e^{-5}$ & 256 \\
H3K9me3 & $1e^{-4}$ & 128 & $2e^{-4}$ & 128 & $1e^{-3}$ & 512 & $1e^{-5}$ & 64 & $1e^{-4}$ & 64 & $1e^{-4}$ & 128 \\
H4K20me1 & $2e^{-5}$ & 128 & $1e^{-3}$ & 256 & $2e^{-4}$ & 64 & $2e^{-5}$ & 128 & $5e^{-5}$ & 256 & $1e^{-5}$ & 64 \\
Enhancer & $2e^{-5}$ & 256 & $2e^{-4}$ & 128 & $5e^{-4}$ & 512 & $5e^{-5}$ & 256 & $5e^{-5}$ & 64 & $2e^{-5}$ & 64 \\
Enhancer types & $5e^{-5}$ & 512 & $1e^{-3}$ & 64 & $2e^{-4}$ & 128 & $2e^{-5}$ & 128 & $2e^{-5}$ & 64 & $5e^{-5}$ & 64 \\
Promoter all & $1e^{-4}$ & 64 & $5e^{-4}$ & 128 & $5e^{-4}$ & 128 & $1e^{-5}$ & 128 & $1e^{-4}$ & 128 & $5e^{-5}$ & 512 \\
Promoter non-TATA & $2e^{-5}$ & 128 & $5e^{-5}$ & 64 & $5e^{-5}$ & 128 & $2e^{-5}$ & 256 & $2e^{-5}$ & 64 & $1e^{-4}$ & 128 \\
Promoter TATA & $2e^{-4}$ & 128 & $5e^{-4}$ & 64 & $1e^{-3}$ & 128 & $5e^{-5}$ & 64 & $5e^{-5}$ & 128 & $1e^{-4}$ & 64 \\
Splice acceptor & $5e^{-5}$ & 256 & $1e^{-3}$ & 128 & $5e^{-3}$ & 64 & $1e^{-5}$ & 256 & $5e^{-5}$ & 128 & $5e^{-5}$ & 64 \\
Splice site all & $5e^{-5}$ & 256 & $1e^{-3}$ & 64 & $2e^{-3}$ & 64 & $1e^{-5}$ & 64 & $1e^{-4}$ & 256 & $1e^{-4}$ & 128 \\
Splice site donors & $2e^{-5}$ & 128 & $2e^{-3}$ & 64 & $5e^{-3}$ & 64 & $2e^{-5}$ & 128 & $5e^{-5}$ & 128 & $2e^{-5}$ & 64 \\
\bottomrule
\end{tabular}
\label{tab:benchmark_hyperparam2}
\end{table*}

\begin{table*}[!htb]
\small
\renewcommand{\arraystretch}{0.8}
\centering
\caption{Hyperparameter settings for the Genomic Benchmarks.}
\resizebox{\textwidth}{!}{%
\begin{tabular}{@{}l*{16}{c}@{}}
\toprule
 & 
\multicolumn{2}{c}{\shortstack{DNABERT-2}} & 
\multicolumn{2}{c}{\shortstack{HyenaDNA}} & 
\multicolumn{2}{c}{\shortstack{NT-v2}} & 
\multicolumn{2}{c}{\shortstack{Caduceus-Ph}} & 
\multicolumn{2}{c}{\shortstack{Caduceus-PS}} & 
\multicolumn{2}{c}{\shortstack{GROVER}} & 
\multicolumn{2}{c}{\shortstack{GENERator}} & 
\multicolumn{2}{c}{\shortstack{GENERator-All}} \\ 
\cmidrule(l){2-3} 
\cmidrule(l){4-5} 
\cmidrule(l){6-7} 
\cmidrule(l){8-9} 
\cmidrule(l){10-11} 
\cmidrule(l){12-13} 
\cmidrule(l){14-15} 
\cmidrule(l){16-17}
 & LR & BS & LR & BS & LR & BS & LR & BS & LR & BS & LR & BS & LR & BS & LR & BS \\ 
\midrule
Coding vs. Intergenomic & $1e^{-5}$ & 64 & $1e^{-4}$ & 64 & $5e^{-5}$ & 512 & $2e^{-4}$ & 128 & $2e^{-4}$ & 128 & $5e^{-5}$ & 256 & $2e^{-5}$ & 256 & $1e^{-5}$ & 128 \\
Drosophila Enhancers Stark & $5e^{-5}$ & 512 & $1e^{-3}$ & 512 & $5e^{-5}$ & 64 & $5e^{-4}$ & 64 & $1e^{-3}$ & 64 & $1e^{-4}$ & 64 & $2e^{-4}$ & 256 & $1e^{-5}$ & 128 \\
Human Enhancers Cohn & $5e^{-5}$ & 128 & $5e^{-5}$ & 128 & $5e^{-5}$ & 256 & $2e^{-4}$ & 64 & $2e^{-4}$ & 256 & $2e^{-5}$ & 256 & $1e^{-5}$ & 128 & $2e^{-5}$ & 64 \\
Human Enhancers Ensembl & $1e^{-4}$ & 256 & $5e^{-4}$ & 512 & $5e^{-5}$ & 128 & $5e^{-4}$ & 256 & $2e^{-4}$ & 64 & $5e^{-5}$ & 512 & $5e^{-5}$ & 128 & $1e^{-4}$ & 512 \\
Human Ensembl Regulatory & $5e^{-5}$ & 128 & $1e^{-4}$ & 128 & $2e^{-5}$ & 128 & $5e^{-4}$ & 64 & $5e^{-4}$ & 64 & $1e^{-5}$ & 128 & $1e^{-5}$ & 128 & $2e^{-5}$ & 512 \\
Human NonTATA Promoters & $2e^{-4}$ & 256 & $1e^{-4}$ & 128 & $5e^{-5}$ & 64 & $5e^{-4}$ & 64 & $5e^{-4}$ & 128 & $5e^{-5}$ & 64 & $5e^{-5}$ & 128 & $1e^{-4}$ & 64 \\
Human OCR Ensembl & $1e^{-4}$ & 256 & $5e^{-5}$ & 256 & $5e^{-5}$ & 128 & $1e^{-3}$ & 256 & $5e^{-4}$ & 128 & $5e^{-5}$ & 64 & $1e^{-5}$ & 64 & $5e^{-5}$ & 128 \\
Human vs. Worm & $2e^{-5}$ & 64 & $1e^{-4}$ & 128 & $2e^{-5}$ & 64 & $2e^{-4}$ & 64 & $2e^{-4}$ & 128 & $2e^{-5}$ & 128 & $2e^{-5}$ & 512 & $2e^{-5}$ & 64 \\
Mouse Enhancers Ensembl & $1e^{-5}$ & 64 & $2e^{-4}$ & 128 & $5e^{-5}$ & 64 & $1e^{-3}$ & 64 & $2e^{-4}$ & 128 & $1e^{-5}$ & 128 & $5e^{-4}$ & 512 & $5e^{-5}$ & 64 \\
\bottomrule
\end{tabular}
}
\label{tab:benchmark_hyperparam3}
\end{table*}

\begin{table*}[!htb]
\small
\renewcommand{\arraystretch}{0.8}
\centering
\caption{Hyperparameter settings for the Gener tasks.}
\resizebox{\textwidth}{!}{%
\begin{tabular}{@{}l*{16}{c}@{}}
\toprule
 & 
\multicolumn{2}{c}{\shortstack{DNABERT-2}} & 
\multicolumn{2}{c}{\shortstack{HyenaDNA}} & 
\multicolumn{2}{c}{\shortstack{NT-v2}} & 
\multicolumn{2}{c}{\shortstack{Caduceus-Ph}} & 
\multicolumn{2}{c}{\shortstack{Caduceus-PS}} & 
\multicolumn{2}{c}{\shortstack{GROVER}} & 
\multicolumn{2}{c}{\shortstack{GENERator}} & 
\multicolumn{2}{c}{\shortstack{GENERator-All}} \\ 
\cmidrule(l){2-3} 
\cmidrule(l){4-5} 
\cmidrule(l){6-7} 
\cmidrule(l){8-9} 
\cmidrule(l){10-11} 
\cmidrule(l){12-13} 
\cmidrule(l){14-15} 
\cmidrule(l){16-17}
 & LR & BS & LR & BS & LR & BS & LR & BS & LR & BS & LR & BS & LR & BS & LR & BS\\ 
\midrule
Gene Classification & $5e^{-5}$ & 64 & $2e^{-4}$ & 128 & $5e^{-5}$ & 128 & $5e^{-4}$ & 128 & $2e^{-4}$ & 64 & $5e^{-5}$ & 128 & $1e^{-5}$ & 64 & $1e^{-4}$ & 64 \\
Taxonomic Classification & $5e^{-5}$ & 64 & $2e^{-4}$ & 64 & $1e^{-4}$ & 512 & $2e^{-4}$ & 128 & $5e^{-4}$ & 256 & $1e^{-4}$ & 128 & $1e^{-5}$ & 128 & $5e^{-6}$ & 64 \\
\bottomrule
\end{tabular}
}
\label{tab:benchmark_hyperparam4}
\end{table*}

\begin{table*}[!htb]
    \small
    \renewcommand{\arraystretch}{1.2}
    \centering
    \caption{Detailed architecture of GENERator.}
    \begin{tabular}{lccc}
        \toprule
        Parameter & GENERator-1B & GENERator-3B \\
        \midrule
        Layers & 26 & 30 \\
        Hidden Size & 2048 & 3072 \\
        Intermediate Size & 5632 & 8448 \\
        Vocabulary Size & 4128 & 4128\\
        Attention Heads & 32 (4 KV heads) & 32 (4 KV heads) \\
        Context Length & 16,384 (98,304 bp) & 16,384 (98,304 bp) \\
        Positional Encoding & RoPE~\cite{rope} & RoPE~\cite{rope} \\
        Hidden Activation & SiLU~\cite{silu} & SiLU~\cite{silu} \\
        \bottomrule
    \end{tabular}
    \label{tab:model_arch}
\end{table*}
\begin{table*}[!htb]
\small
\renewcommand{\arraystretch}{1.2}
\centering
\caption{Pre-training data statistics (taxonomic group).}
\begin{tabular}{lccc}
\toprule
Taxonomic group & Number of genes & Number of nucleotides (bp) \\
\midrule
Protozoa & 1,107,499 & 2,034,179,213 \\
Fungi & 6,299,990 & 10,593,031,563 \\
Plant & 7,279,591 & 30,548,451,231 \\
Invertebrate & 7,602,349 & 71,979,748,208 \\
Vertebrate (other) & 10,915,710 & 161,686,774,317 \\
Mammalian & 6,837,553 & 109,406,723,311 \\
\bottomrule
\end{tabular}
\label{tab:pretrain_data_statistics_species}
\end{table*}

\begin{table*}[!htb]
\small
\renewcommand{\arraystretch}{1.2}
\centering
\caption{Pre-training data statistics (gene type).}
\begin{tabular}{lccc}
\toprule
Gene type & Number of genes & Number of nucleotides (bp) \\
\midrule
Protein Coding & 30,883,451 & 340,466,000,000 \\
miscRNA & 342,761 & 6,448,121,770 \\
ncRNA & 4,151,851 & 28,538,620,565 \\
pseudo & 2,322,278 & 10,307,467,170 \\
rRNA & 654,807 & 360,329,916 \\
tRNA & 1,687,523 & 128,397,507 \\
tmRNA & 21 & 9,793 \\
\bottomrule
\end{tabular}
\label{tab:pretrain_data_statistics_gene}
\end{table*}
\begin{table*}[!htb]
\small
\renewcommand{\arraystretch}{1.2}
\centering
\caption{Summary of baseline models.}
\resizebox{\textwidth}{!}{
\begin{tabular}{lccccccc}
\toprule
Model Name & Pre-train Task & Architecture & Tokenizer & Pre-train Data Scope & Pre-train Data Volume (bp) & Context Length (bp) & Parameter Size \\
\midrule
Enformer & Supervised & CNN+Encoder & Single Nucleotide & Human, Mouse & 64B & 200K & 252M \\
DNABERT-2 & MLM & Encoder & BPE & Multispecies & 32.5B & 3K & 117M \\
HyenaDNA & NTP & Hyena & Single Nucleotide & Human & 3B & 1M & 55M \\
NT & MLM & Encoder & 6-mer & Multispecies & 174B & 6K & 2.5B \\
NT-v2 & MLM & Encoder & 6-mer & Multispecies & 174B & 12K & 500M \\
Caduceus & MLM & BiMamba & Single Nucleotide & Human & 35B & 131K & 8M \\
GROVER & MLM & Encoder & BPE & Human & 3B & 3K & 87M \\
GENERator-1B & NTP & Decoder & 6-mer & Multispecies & 386B & 98K & 1.2B \\
GENERator-3B & NTP & Decoder & 6-mer & Multispecies & 386B & 98K & 3B \\
GENERator-All & NTP & Decoder & 6-mer & Multispecies & 1.9T & 98K & 1.2B \\
Evo2-1B-Base & NTP & StripedHyena 2 & Single Nucleotide & Multispecies & 1T & 8K & 1.1B \\
Evo2-7B-Base & NTP & StripedHyena 2 & Single Nucleotide & Multispecies & 2.1T & 8K & 6.5B \\
Evo2-7B & NTP & StripedHyena 2 & Single Nucleotide & Multispecies & 2.4T & 1M & 6.5B \\
\bottomrule
\end{tabular}
}
\label{tab:baseline_introduction}
\end{table*}
\begin{table*}[!htb]
\small
\renewcommand{\arraystretch}{1.2}
\centering
\caption{Computational resource usage ({\color{gray}gray color denotes that the model was trained for less than 1 epoch}).}
\begin{tabular}{llrc}
\hline
 & Model & Computational Resources & Data-parallel Size \\ \hline
\multirow{15}{*}{Development} & BPE-512 & 11,551 V100 hours & 32 \\
& BPE-1024 & 10,562 V100 hours & 32 \\
& BPE-2048 & 8,431 V100 hours & 32 \\
& BPE-4096 & 11,155 V100 hours & 64 \\
& BPE-8192 & 8,118 V100 hours & 32 \\
& 2-mer & 28,380 V100 hours & 64 \\
& 3-mer & 15,270 V100 hours & 32 \\
& 4-mer & 11,635 V100 hours & 32 \\
& 5-mer & 9,202 V100 hours & 32 \\
& 6-mer & 9,494 V100 hours & 64 \\
& 7-mer & 8,133 V100 hours & 64 \\
& 8-mer & 7,883 V100 hours & 64 \\
& {\color{gray}Single Nucleotide (Llama)} & 1,350 A100 hours & 8 \\
& {\color{gray}Single Nucleotide (Mamba)} & 1,465 A100 hours & 8 \\
\hline
\multirow{2}{*}{Pre-train}
& GENERator-1B & 11,793 A100 hours & 32 \\
& GENERator-3B & 29,625 A100 hours & 32 \\
& GENERator-All & 9,494 A100 hours & 32 \\ \hline
Downstream & 46 runs per task per model & $>$ 64,000 V100 hours & 8 \\ \hline
\end{tabular}
\label{tab:computational_resources}
\end{table*}

\end{document}